\documentclass[lettersize,journal]{IEEEtran}

\usepackage{algorithmic}
\usepackage{array}
\usepackage{url}
\usepackage{verbatim}
\usepackage{cite}
\usepackage{amsmath,amssymb,amsfonts}
\usepackage[table,xcdraw]{xcolor}

\usepackage{algorithm}
\usepackage{graphicx}
\usepackage{textcomp}
\usepackage{xcolor}
\usepackage{adjustbox}
\usepackage{multirow}

\usepackage{subfigure}
\usepackage{caption}
\usepackage{enumerate}
\usepackage{booktabs} 
\usepackage{pifont}

\usepackage{quoting}

\newcommand{\cmark}{\ding{51}}%
\newcommand{\xmark}{\ding{55}}%

\usepackage{amsmath,amsfonts,bm}

\def\eqref#1{Eq.~(\ref{#1})}

\def\1{\bm{1}}

\def\vm{{\bm{m}}}

\def\vx{{\bm{x}}}

\DeclareMathAlphabet{\mathsfit}{\encodingdefault}{\sfdefault}{m}{sl}
\SetMathAlphabet{\mathsfit}{bold}{\encodingdefault}{\sfdefault}{bx}{n}

\def\gD{{\mathcal{D}}}

\def\gX{{\mathcal{X}}}
\def\gY{{\mathcal{Y}}}

\newcommand{\E}{\mathbb{E}}
\newcommand{\Ls}{\mathcal{L}}

\def\BibTeX{{\rm B\kern-.05em{\sc i\kern-.025em b}\kern-.08em
    T\kern-.1667em\lower.7ex\hbox{E}\kern-.125emX}}
\usepackage{balance}

\begin{document}

\title{Expose Before You Defend: Unifying and Enhancing Backdoor Defenses via Exposed Models}

\author{Yige Li, Hanxun Huang, Jiaming Zhang, Xingjun Ma, and Yu-Gang Jiang,~\IEEEmembership{Fellow, ~IEEE}
\thanks{Yige Li, Xingjun Ma, and Yu-Gang Jiang are with the Shanghai Key Lab of Intell. Info. Processing, School of CS, Fudan University, Shanghai, China (e-mail: xdliyige@gmail.com, xingjunma@fudan.edu.cn, ygj@fudan.edu.cn). }%
    \thanks{Hanxun Huang is with the School of Computing and Information Systems, the University of Melbourne, Australia (e-mail: hanxun@unimelb.edu.au).}%
    \thanks{Jiaming Zhang is with the Department of Computer Science and Engineering, Hong Kong University of Science and Technology, Hong Kong, China (e-mail:jmzhang@ust.hk)}
    \thanks{Work done during Yige's internship at Fudan University.} 
    \thanks{Corresponding Author: Xingjun Ma}
}


\maketitle

\begin{abstract}
Backdoor attacks covertly implant triggers into deep neural networks (DNNs) by poisoning a small portion of the training data with pre-designed backdoor triggers. This vulnerability is exacerbated in the era of large models, where extensive (pre-)training on web-crawled datasets is susceptible to compromise.
In this paper, we introduce a novel two-step defense framework named \emph{Expose Before You Defend (EBYD)}. EBYD unifies existing backdoor defense methods into a comprehensive defense system with enhanced performance. Specifically, EBYD first exposes the backdoor functionality in the backdoored model through a model preprocessing step called \emph{backdoor exposure}, and then applies detection and removal methods to the exposed model to identify and eliminate the backdoor features.
In the first step of backdoor exposure, we propose a novel technique called \textbf{Clean Unlearning (CUL)}, which proactively unlearns clean features from the backdoored model to reveal the hidden backdoor features. We also explore various model editing/modification techniques for backdoor exposure, including fine-tuning, model sparsification, and weight perturbation.
Using EBYD, we conduct extensive experiments on 10 image attacks and 6 text attacks across 2 vision datasets (CIFAR-10 and an ImageNet subset) and 4 language datasets (SST-2, IMDB, Twitter, and AG’s News). The results demonstrate the importance of backdoor exposure for backdoor defense, showing that the exposed models can significantly benefit a range of downstream defense tasks, including backdoor label detection, backdoor trigger recovery, backdoor model detection, and backdoor removal.
More importantly, with backdoor exposure, our EBYD framework can effectively integrate existing backdoor defense methods into a comprehensive and unified defense system. 
We hope our work could inspire more research in developing advanced defense frameworks with exposed models.
Our code is available at \url{https://github.com/bboylyg/Expose-Before-You-Defend}.

\end{abstract}


\section{Introduction}

Deep neural networks (DNNs) trained on large-scale datasets have demonstrated remarkable performance in addressing complex real-world problems across various domains, including computer vision (CV) \cite{he2016deep,dosovitskiy2020image} and natural language processing (NLP) \cite{devlin2018bert,brown2020language}. However, recent studies have shown that DNNs are vulnerable to backdoor attacks \cite{huang2023distilling,li2024multi}, which insert malicious triggers into the model parameters to compromise its test-time predictions.
Specifically, these attacks establish a covert correlation between a predefined trigger pattern and an adversary-specified target label by poisoning a small subset of the training data.
A backdoored model maintains normal performance on clean inputs but consistently misclassifies inputs containing the trigger pattern to the target label.
Importantly, backdoor attacks are not limited to a specific domain; they can compromise both vision and language models. For instance, in the image domain, attackers may manipulate a few pixels or embed specific patterns, while in the text domain, they might incorporate particular words or syntactic structures to trigger malicious behavior.
With the proliferation and accessibility of pre-trained vision and language models from platforms like Hugging Face \cite{wolf2019huggingface}, ensuring the secure and backdoor-free deployment of these models in downstream applications has become increasingly critical.

Existing defense methods against backdoor attacks can be broadly categorized into two types: \emph{detection methods} and \emph{removal methods}. Detection methods identify the existence of a backdoor attack (i.e., trigger) in a trained model (a task known as \emph{backdoor model detection}) or in a training/test sample (a task known as \emph{backdoor sample detection}). Both tasks involve inverting the trigger pattern used by the attack and identifying the targeted class of the attacker \cite{wang2019neural,liu2019abs,huang2023distilling}. Arguably, the ultimate goal of backdoor defense is to completely eliminate the backdoor trigger from a compromised model. This objective lies at the core of backdoor removal methods using techniques such as fine-tuning, pruning \cite{liu2018fine,li2023reconstructive}, or knowledge distillation \cite{li2021neural}. 

While both backdoor detection and removal methods have shown promising results, they have been applied independently, without benefiting from each other. For example, trigger inversion methods often struggle to identify the backdoor class and thus have to assume it is known to the defender, while backdoor removal methods cannot pinpoint the exact trigger pattern and backdoor class. Moreover, both types of methods exhibit performance limitations against several advanced attacks. To date, a unified defense framework capable of effectively detecting and removing all types of backdoor attacks remains absent from the current literature. Additionally, none of the existing defense techniques have demonstrated effectiveness against both image and text backdoor attacks.

\begin{quoting}
    \begin{center}
        \textit{``A known enemy is easier to defeat."\\ \textemdash Ancient Wisdom}
    \end{center}
\end{quoting}

In this work, we aim to address the limitations of existing defenses by drawing inspiration from the ancient wisdom: ``A known enemy is easier to defeat". Intuitively, if we could expose the backdoor within a compromised model through a specialized model preprocessing/editing technique that isolates the backdoor functionality, the backdoor trigger would become much easier to detect, recover, and remove. This could potentially lead to a holistic defense framework against the backdoors. 
This approach is feasible due to the inherent nature of backdoors: the backdoor functionality injected into the victim model is specifically designed to be independent of its normal functionality (to avoid impacting the clean performance). This motivates us to propose the \textbf{\textit{Expose Before You Defend (EBYD)}} framework. EBYD consists of two steps: 1) \emph{backdoor exposure}, a preprocessing step that reveals the backdoor functionality in the compromised model, and 2) \emph{backdoor defense}, which applies existing detection and removal techniques to the preprocessed (exposed) model to enhance overall performance.

In EBYD, \emph{backdoor exposure} plays a crucial role in connecting and enhancing different defense techniques. However, decoupling and exposing the backdoor functionality from a compromised model is a challenging task, as evidenced by the shortcomings of current defense methods \cite{tan2020bypassing,qi2022revisiting}. To address this, we propose a novel technique called \textbf{Clean Unlearning (CUL)}, which exposes backdoor functionality by unlearning the clean functionality from the backdoored model rather than directly searching for backdoor features. Intuitively, a model can be effectively unlearned by maximizing its error on a few clean samples. Although this type of lightweight unlearning may be partial, it is sufficient to inhibit the clean functionality of the model for the purpose of backdoor exposure. 
Following this, we conduct a comprehensive exploration of possible model preprocessing techniques, including fine-tuning, model sparsification, and weight perturbation. We demonstrate that these techniques can also expose the backdoor functionality in a compromised model.

In our EBYD framework, the exposed model provides a better starting point for all subsequent defenses. It not only enhances existing backdoor removal methods but also unifies various backdoor defense tasks, including trigger inversion, backdoor label detection, and backdoor sample detection.
For instance, when combined with Neural Cleanse (NC) \cite{wang2019neural}, one of the most effective methods for trigger inversion and backdoor model detection, EBYD not only improves NC's detection rate but also facilitates the identification of the backdoor label (class). Similarly, when integrated with STRIP \cite{gao2019strip}, a well-established method for backdoor sample detection, EBYD enables the detection of backdoor samples that are significantly more complex and stealthy than traditional attacks. Moreover, the backdoor-exposed model enhances the effectiveness of existing backdoor removal methods \cite{liu2018fine,zheng2022data,li2023reconstructive}, elevating their performance to a higher level.

More importantly, we demonstrate that EBYD can be extended to language models to defend against a wide range of textual backdoor attacks. As such, EBYD serves as a unifying framework that integrates various defense methods,  enabling independent strategies like backdoor detection, trigger recovery, and backdoor removal to collaborate and contribute to a comprehensive defense system.
With EBYD, we conduct the most extensive defense evaluation to date, defending against 10 image attacks and 6 text attacks. 
Empirical results across two image datasets (CIFAR-10 and an ImageNet subset) and four text datasets (SST-2, IMDB, Twitter, and AG’s News), employing various model architectures, demonstrate that our EBYD defense framework achieves significant performance improvements over current state-of-the-art (SOTA) methods.

In summary, the main contributions of this work are:
\begin{itemize}
\item We introduce a defense framework named \emph{Expose Before You Defend (EBYD)} that decouples backdoor defense into two steps. The first step, \emph{backdoor exposure}, exposes the backdoor functionality contained in the model, while the second step focuses on detecting and removing the backdoor functionality.

\item We propose a novel backdoor exposure technique named \textbf{Clean Unlearning (CUL)} which unlearns the clean features from the model to expose the backdoor functionality. We demonstrate that CUL remains effective even when unlearning is performed on a few clean samples.
The unlearned model provides a good starting point to unify detection and removal defenses.

\item Under EBYD, we first explore various model preprocessing techniques for backdoor exposure, based on which we have conducted the most comprehensive empirical evaluation in the field involving both visual and language backdoor attacks. Our results demonstrate the effectiveness and universality of our EBYD against 16 types of backdoor attacks (10 image attacks and 6 textual attacks).

\end{itemize}

This work is an extension of our conference paper \cite{li2023reconstructive} presented at the Fortieth International Conference on Machine Learning (ICML), 2023. We have made the following major extensions:

\begin{enumerate}
    \item We have extended the clean unlearning technique introduced in our conference paper into a more general module, \emph{Backdoor Exposure}, and building on this, we introduced a new and systematic two-step backdoor defense framework: \emph{Expose Before You Defend (EBYD)}.

    \item We have conducted a comprehensive exploration of potential model exposure techniques not covered in the conference paper. These include model-level techniques (pruning and parameter adversarial perturbation) and data-level techniques (unlearning and fine-tuning).

    \item  We have extended our defense experiments to the text domain, enabling the first-ever evaluation of backdoor defense methods across both vision and language tasks. 
\end{enumerate}

\section{Related Work}

\subsection{Backdoor Attack} 
A backdoor attack aims to implant a malicious trigger into the victim models at training time by poisoning a small proportion of the training samples with a carefully crafted trigger pattern. After training on the poisoned data, the trigger pattern becomes strongly correlated with the backdoor target class. Depending on the adversary's capabilities and design of the trigger pattern, existing backdoor attacks can be broadly categorized into \textbf{data-poisoning attacks} and \textbf{training-manipulation attacks}. In data-poisoning attacks, the adversary injects a pre-defined trigger pattern into a small proportion of the training data to trick the model into learning the connection between the trigger pattern and a backdoor label \cite{wu2022backdoorbench}. The trigger pattern can be relatively simple, such as a single pixel \cite{tran2018spectral}, a black-and-white square \cite{gu2017badnets}, random noise \cite{chen2017targeted} or more complex patterns such as adversarial perturbation \cite{turner2019clean}, and input-aware patterns \cite{nguyen2020input}. On the other hand, training-manipulation attacks directly manipulate the training procedure to optimize for the backdoor objective in the feature space, using techniques such as feature collision \cite{shafahi2018poison} or by directly modifying model parameters via weight perturbation \cite{garg2020can}.

Additionally, textual backdoor attacks leverage training data poisoning with various types of triggers. These include rare or meaningless words, such as ‘cf’ \cite{chen2021badnl}, and syntactic structure manipulation \cite{qi2021hidden}. More recent approaches aim to design sophisticated triggers using techniques like layer-wise poisoning \cite{li2021layerwise} and constrained optimization \cite{yang2021rethinking}, enhancing both attack effectiveness and stealthiness. All of these methods have demonstrated significant success and continue to challenge existing defense mechanisms.

\begin{table}[!tp]
\caption{Functionalities of existing backdoor defenses: backdoor exposure (BE), backdoor model detection (BMD), backdoor sample detection (BSD) or backdoor removal (BR). }
\centering
\small
\begin{sc}
    \begin{tabular}{ccccc}
    \toprule
    Defense Method & BE & BMD & BSD & BR  \\ \hline
    NC & \xmark & \cmark & \xmark & \xmark\\
    STRIP & \xmark & \xmark & \cmark & \xmark\\
    Fine-pruning & \xmark & \xmark & \xmark & \cmark\\
    ABL & \xmark & \xmark & \cmark & \cmark\\
    I-BAU & \xmark & \xmark & \xmark & \cmark\\
    ANP & \xmark & \xmark & \xmark & \cmark\\
    RNP & \cmark & \xmark & \xmark & \cmark\\
    EBYD (Ours) & \cmark & \cmark & \cmark & \cmark\\ \bottomrule
    \end{tabular}
\end{sc}
\vskip -0.1in
\label{tab:functions}
\end{table}

\subsection{Backdoor Defense} 
Numerous approaches have been proposed to defend DNNs against backdoor attacks, among which backdoor detection and backdoor removal methods are the two most prevalent strategies. 

\textbf{Backdoor Detection.} 
Several detection methods identify backdoors based on the prediction bias observed in different input examples \cite{li2020rethinking} or the statistical deviation in the feature space \cite{tran2018spectral, chen2018detecting}.  
More effective detection methods leverage reverse engineering techniques to recover the trigger pattern and then identify the backdoor label by anomaly detection \cite{wang2019neural,liu2019abs}. One representative method is Neural Cleanse (NC) \cite{wang2019neural}, which recovers trigger patterns that can alter the model's predictions with minimum perturbation. Other methods focus on detecting backdoored samples at inference time, such as the STRIP method \cite{gao2019strip}. Numerous detection methods have been proposed in the NLP domain to identify potential trigger words by analyzing their influence on model outputs \cite{BKI,RAP}.

\textbf{Backdoor Removal.} Backdoor removal methods aim to erase backdoors from compromised models without significantly degrading their performance on clean samples. This line of work includes Fine-tuning, Fine-pruning \cite{liu2018fine}, Mode Connectivity Repair \cite{zhao2020bridging}, and Neural Attention Distillation (NAD) \cite{li2021neural}.  More recently, a training-time defense method called Anti-Backdoor Learning (ABL) \cite{li2021anti} has been proposed to train clean models directly on backdoored data. Meanwhile, Adversarial Unlearning of Backdoors via Implicit Hypergradient (I-BAU) \cite{zeng2021adversarial} is proposed to cleanse backdoored model with adversarial training. Adversarial Neuron Pruning (ANP) \cite{wu2021adversarial} prunes neurons that are more sensitive to adversarial perturbations to remove backdoors. The latest method, Reconstructive Neuron Pruning (RNP), has set a new state-of-the-art in defending against data-poisoning backdoor attacks \cite{li2023reconstructive}. The study conducted in RNP \cite{li2023reconstructive} shows that one can reveal backdoor-related features (neurons) by unlearning the model on a small portion of clean data.  In NLP defense, the MF approach \cite{MF} mitigates backdoor learning by minimizing overfitting but struggles with attacks involving textual styles and grammatical patterns. Additionally, CUBE suggests that clustering in the feature space can help identify and remove backdoor samples, although this might impact the accuracy of clean tasks \cite{Openbackdoor}. However, these methods lack generalizability and struggle to precisely expose the underlying backdoor behaviors, particularly the hidden triggers in language models. How to effectively reveal backdoor behaviors hidden in the language models is an open research problem that deserves more exploration. Table \ref{tab:functions} summarizes the functionalities of existing and our proposed EBYD defense methods.

\begin{figure*}[!tp]
\centering
\includegraphics[width=0.92\linewidth]{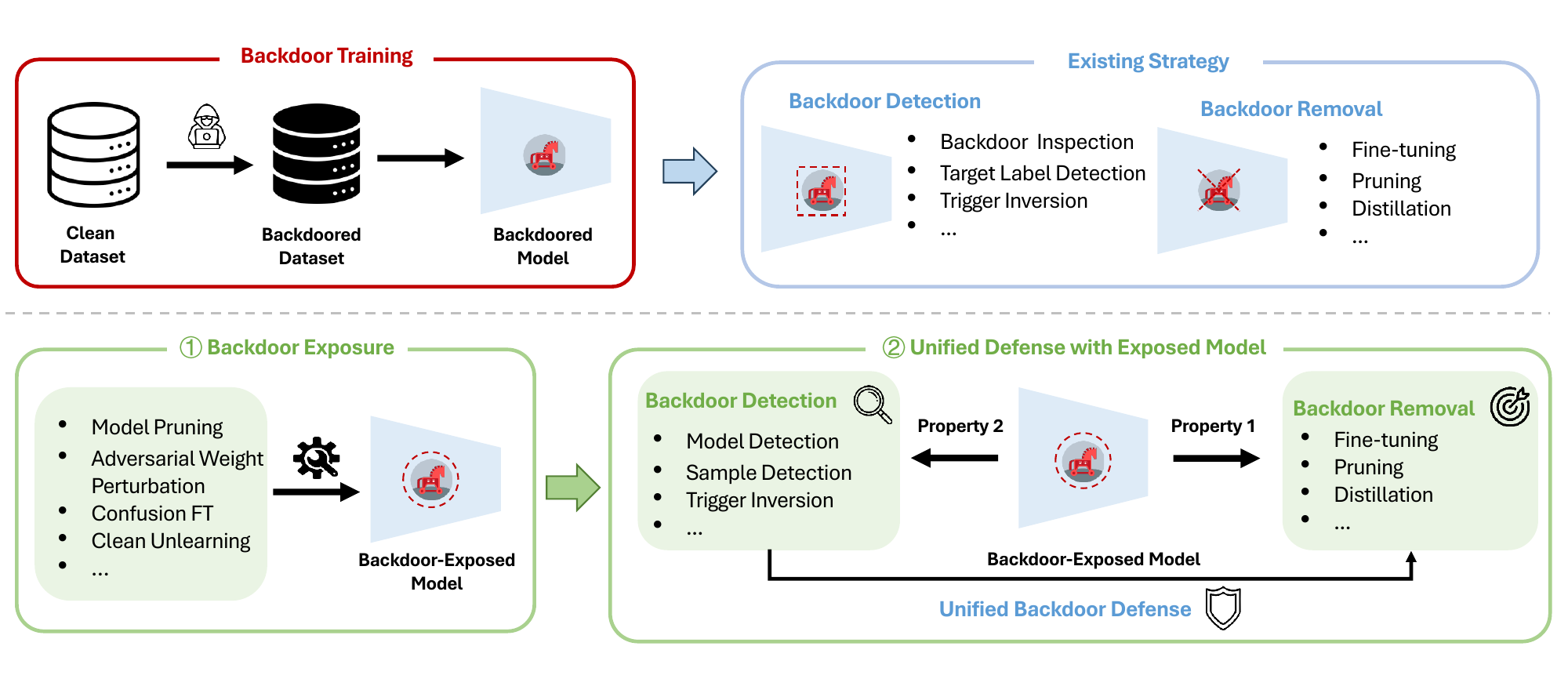}
\vskip -0.1in
\caption{\textbf{Top}: Traditional backdoor defense pipeline; \textbf{Bottom}: Our proposed two-step defense framework EBYD.}
\label{fig:EBYD_overview}
\end{figure*}

\subsection{Understandings of Backdoors}
A set of understandings and assumptions regarding backdoors has developed during the process of backdoor attack and defense. We summarize these assumptions and highlight they are necessary for successful backdoor defense.

\textbf{Backdoor attack creates shortcuts in DNNs.} 
The distinctive behavior of the backdoored model on clean versus backdoor samples indicates the existence of neural shortcuts~\cite{liu2019abs, geirhos2020shortcut} in backdoored models. These shortcuts have been found to be learned at an early stage of training at a much faster rate than normal features~\cite{li2021anti}.
As a result, defenders can leverage this shortcut behavior to determine whether a model has been backdoored. One such method is the Neural Cleanse (NC) \cite{wang2019neural} which detects a backdoored model by searching a shortcut modification (i.e., the trigger pattern) of an arbitrary input toward a backdoor target label. This works reasonably well against attacks like BadNets \cite{gu2017badnets}, Blend \cite{chen2017targeted}, and Trojan \cite{liu2018trojaning}. However, revealing shortcuts becomes increasingly challenging for complex and dynamic attacks, such as sample-wise dynamic attacks~\cite{nguyen2020input} and WaNet~\cite{nguyen2021wanet}. In this case, simple shortcut discovery techniques like NC tend to fail as experimented in several existing works~\cite{liu2020reflection,nguyen2020input,cheng2021deep,nguyen2021wanet}.

\textbf{Backdoor samples have anomaly output distributions.} 
This understanding was established with the success of backdoor sample detection methods like STRIP~\cite{gao2019strip}. The distinguishable differences in output distributions between clean and backdoor samples can be statistically characterized to build accurate detectors against simple backdoor attacks like BadNets \cite{gu2017badnets}, Blend \cite{chen2017targeted}, and Trojan \cite{liu2018trojaning}. For instance, STRIP detects potential backdoor samples based on the relative entropy of the output distribution. However, such statistical differences can be easily suppressed by adaptive attacks~\cite{shafahi2018poison,nguyen2020input}, leading to detection failures.

\textbf{Backdoor features are only activated by backdoor triggers.}
It has been observed that some neurons are hibernating on clean samples and can only be activated by the trigger pattern~\cite{liu2018fine}. These neurons are referred to as backdoor neurons and can potentially be identified as those less useful for normal classification (i.e., less activated by clean samples). However, recent works have shown that the Fine-pruning defense suffers from severe accuracy degradation when only small clean data are available~\cite{wu2021adversarial} and is ineffective against adaptive attacks \cite{yao2019latent,nguyen2020input,nguyen2021wanet}.
Arguably, the failure of Fine-pruning is caused by an inaccurate decoupling of the backdoor functionality/features. In this work, we address this issue against a wide range of advanced attacks through an independent backdoor exposure step.

The effectiveness of existing backdoor detection and removal methods often depends on the assumptions of the distinctive behavior between the backdoor and clean functionalities embedded in the backdoored model. In this paper, we propose a simple yet versatile defense framework based on the insight of ``backdoor exposure." We demonstrate that an exposed (preprocessed) model using different techniques can significantly boost the performance of existing backdoor detection and backdoor removal methods. Moreover, the exposed model enables us to, for the first time in the literature, integrate detection, trigger inversion, and removal methods into a cohesive defense pipeline.

\section{Proposed EBYD Framework}
In this section, we start by describing the threat model, followed by a brief overview of our EBYD framework. We introduce the key steps of EBYD in the next two sections.

\subsection{Threat Model} 
The threat model adopted in this work encompasses three common backdoor scenarios: \emph{untrusted datasets}, \emph{model outsourcing}, and \emph{pre-trained models}. In model outsourcing, developers may use third-party platforms, such as Machine Learning as a Service (MLaaS) \cite{ribeiro2015mlaas}, due to limited technical capabilities or computational resources. Malicious attackers can exploit these platforms to manipulate training data, embedding backdoors into the model during training. This scenario is particularly vulnerable because attackers have full access to the training data, model, triggers, and training process, after which the compromised model is returned to the developers.
Another attack vector involves pre-trained models \cite{li2021neural}. Attackers may release pre-trained models with embedded backdoor triggers on model repositories (e.g., Hugging Face or GitHub). Victims may unknowingly download these models and use them for downstream tasks via transfer learning. Additionally, attackers might first infect a popular pre-trained model with a backdoor and then redistribute the modified model to repositories.

For backdoor defense, we assume that the defender has full access to the victim (potentially backdoored) model and a small set of clean data (approximately 1\%) as defense data $\gD_d$ for backdoor exposure, model detection, or trigger removal. The defense data is assumed to be independent and identically distributed (i.i.d.) with the training and test data, which is a standard assumption in existing defenses.

\subsection{Framework Overview}
As illustrated in Fig. \ref{fig:EBYD_overview}, our proposed EBYD is a two-step defense framework that first leverages a backdoor exposure method to reveal the backdoor functionality hidden in the model and then applies a detection or removal method to identify the backdoor class, reverse engineer the trigger, and finally remove the backdoor from the model. 
The detailed defense objectives of each step are outlined as follows:

\begin{itemize}
    \item \textbf{Backdoor Exposure.}
    Given an unknown deep model (whether it's backdoored or clean), we leverage an exposure technique to unveil the model's potential (backdoor) characteristics. If the model contains a backdoor, the objective is to obtain a backdoor-exposed model that includes nearly all backdoor-related features while eliminating the functionality of clean features. This backdoor-exposed model serves as valuable prior information for downstream tasks such as backdoor detection and removal.

    \item \textbf{Unified Defense with Exposed Model.}
  This step achieves two defense objectives: backdoor detection and backdoor removal. For backdoor detection, we propose leveraging the exposed model generated by the aforementioned exposure techniques to determine the presence of a backdoor. For backdoor removal, we restore the clean performance of the backdoor-exposed model and eliminate backdoor behavior using our proposed \emph{Recover-Pruning} method. \\

\end{itemize}

Our EBYD framework serves as a unified pipeline that integrates different types of defense methods, enabling independent strategies such as backdoor model detection, backdoor sample detection, and backdoor removal to work collaboratively toward a comprehensive defense system.

\section{Backdoor Exposure}
In this section, we introduce our proposed backdoor exposure method, \emph{Clean Unlearning (CUL)}, and several alternative techniques we explored in this paper. We then discuss the implications of each technique for uncovering the backdoor functionalities.

\subsection{Clean Unlearning} \label{sec:clean unlearning}
Taking image classification task as an example, let $\gD = \{(\vx_i, y_i)\}_{i=1}^n$ represent the original training dataset, where $\vx_i \in \gX$ represents a clean training image and $y_i \in \gY$ is its true label. The goal of a backdoor attack is to add a specific pattern or perturbation as the backdoor trigger $\Delta$ on the original input sample $\vx$. The construction process of the triggered sample $\vx_b$ can be represented as:
$\vx_b= \vx \odot(1-\boldsymbol{m})+\Delta \odot \boldsymbol{m}$, where $\odot$ denotes element-wise multiplication, and $\boldsymbol{m}$ represents a non-zero image mask that controls the region where the trigger is added.

Once the backdoor triggers are implanted into the clean samples, the backdoored dataset can be represented as $\hat{\gD} = \gD_c \cup \gD_b$, where $\gD_c = {(\vx_c, y_c)}$ represents clean samples and their original labels, and $\gD_b = {(\vx_b, y_b)}$ represents triggered samples and their backdoor targeted labels. Training a backdoored model on $\hat{\gD}$ can be formalized as:
\begin{equation} \label{eq:1}
\begin{aligned}
\underset{\theta=\theta_{c}\cup\theta_{b}}{\arg \min } & \Big[ \underbrace{ \mathbb{E}_{(\boldsymbol{x}_c, y_c)\in \mathcal{D}_{c}}\mathcal{L}(f(\boldsymbol{x}_c,\ y_{c};\ \theta_{c}))}_{\text{clean task}} \\
& + \underbrace{ \mathbb{E}_{(\boldsymbol{x}_b, y_b)\in \mathcal{D}_{b}} \mathcal{L}(f(\boldsymbol{x}_b,\ y_{b};\ \theta_{b})) }_{\text{backdoor task}} \Big],
\end{aligned}
\end{equation}
where $\mathcal{L}$ is the classification loss (e.g., cross-entropy). Backdoor learning can be viewed as a dual-task learning process that simultaneously optimizes the clean and backdoor tasks. Note that, although $\theta=\theta_{c}\cup\theta_{b}$, it does not mean $\theta_{c}$ cannot overlap with $\theta_{b}$, i.e., it is possible that $\theta_{c}\cap\theta_{b}\neq \emptyset$. 

Given a backdoored model $f(\cdot\,; \theta_c \cup \theta_b)$, the goal of backdoor exposure is to reveal the backdoor functionality via an exposure function $\Phi$:
\begin{equation}\label{eq:cul}
\Phi: f\left(\cdot\, ;\, \theta_c \cup \theta_b \right) \rightarrow f\left(\cdot\, ; \, \theta_b \right).
\end{equation}

Since the defender does not know the poisoned samples, directly exposing the neurons associated with the backdoor functionality—referred to as \emph{backdoor neurons}—is infeasible. However, the defender possesses a small set of clean samples, termed \emph{defense data} in our threat model, which can be used to defend the model. This leads us to approach backdoor exposure by suppressing or erasing the clean neurons identified by the defense data. Specifically, we design exposure strategies to maximize the model's classification loss on the clean parameters $\theta_c$ while preserving the backdoor functionality on the backdoor parameters $\theta_b$.

To achieve this, we introduce a simple yet effective backdoor exposure technique called Clean Unlearning (CUL), which unlearns the clean features from the backdoored model to reveal the backdoor features. Our CUL method focuses on unlearning the model using specifically designed defense data. Intuitively, the clean features (or clean performance) can be unlearned regarding a particular task by maximizing its loss on data defining that task, which is the inverse of the training process. This approach leads us to solve the following maximization problem:
\begin{equation} \label{eq:EBYD_max}
\begin{aligned}
\max_{\theta_c} \E_{(\boldsymbol{x}_d,\ y_{d}) \in \mathcal{D}_{d}} \|\mathcal{L} (f\left(\boldsymbol{x}_d,\ y_{d} ; \theta_c \cup \theta_b \right)) - \gamma\|, 
\end{aligned}
\end{equation}
where $\Ls$ is the cross-entropy loss, $\|\cdot\|$ denotes the absolute operator, $(\vx_d,y_d) \in \gD_d$ are the clean defense samples, and $\gamma$ is a pre-defined threshold used to prevent loss explosion due to gradient ascent.

The CUL method defined in \eqref{eq:EBYD_max} enables the model to unlearn the functionality defined by the samples in dataset $\mathcal{D}_{d}$. In a backdoored model, this unlearning process forces the model to forget general clean features (e.g., `cat' or `dog') while preserving the backdoor-associated features. This is because backdoor attacks are often designed to be independent of the clean functionality, minimizing their impact on the model's clean performance to remain stealthy.
More importantly, clean unlearning can be achieved very efficiently on a few clean samples. \\

\begin{figure*}[!tp]
\vskip 0.1in
\centering
    \subfigure[The exposure effectiveness (w.r.t. \textbf{Property 1}) for 3 backdoored models. ]{\label{fig:prediction_prop}
        \includegraphics[width=0.23\linewidth]{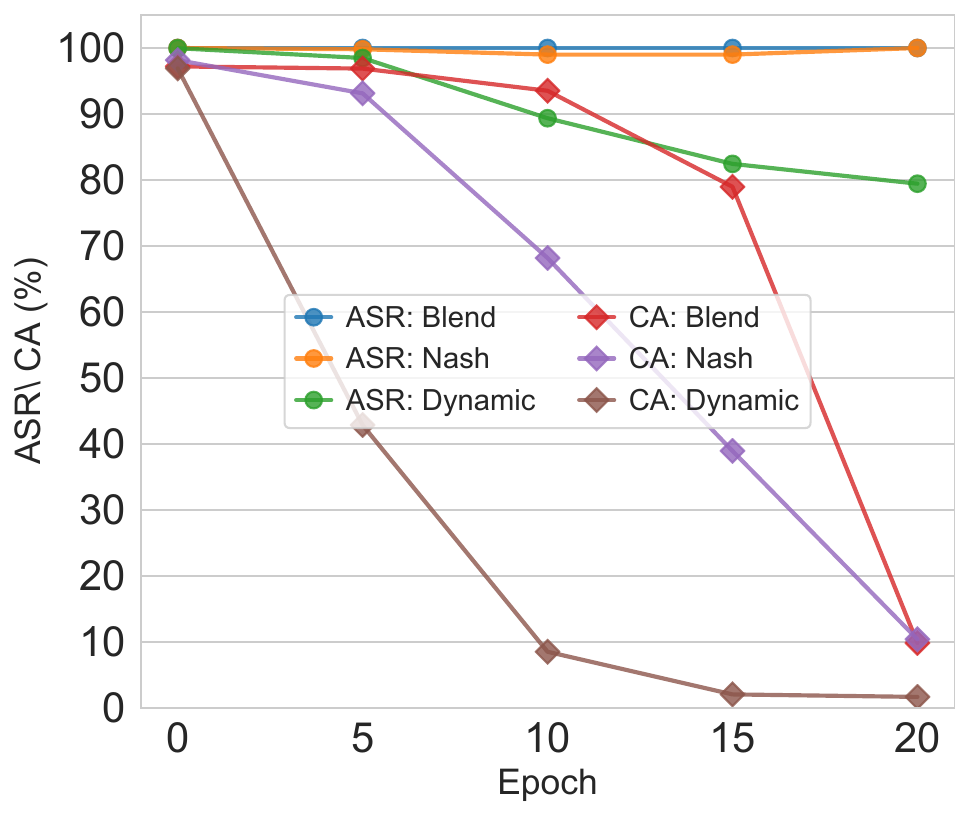}
        }\hfill
    \subfigure[The prediction consistency (w.r.t. \textbf{Property 2}) of each class for BadNets (\textit{Left}), Nash (\textit{Middle}), and Dynamic attacks (\textit{Right}) respectively under Clean Unlearning (CUL).]{\label{fig:classes_prop}
        \includegraphics[width=0.23\linewidth]{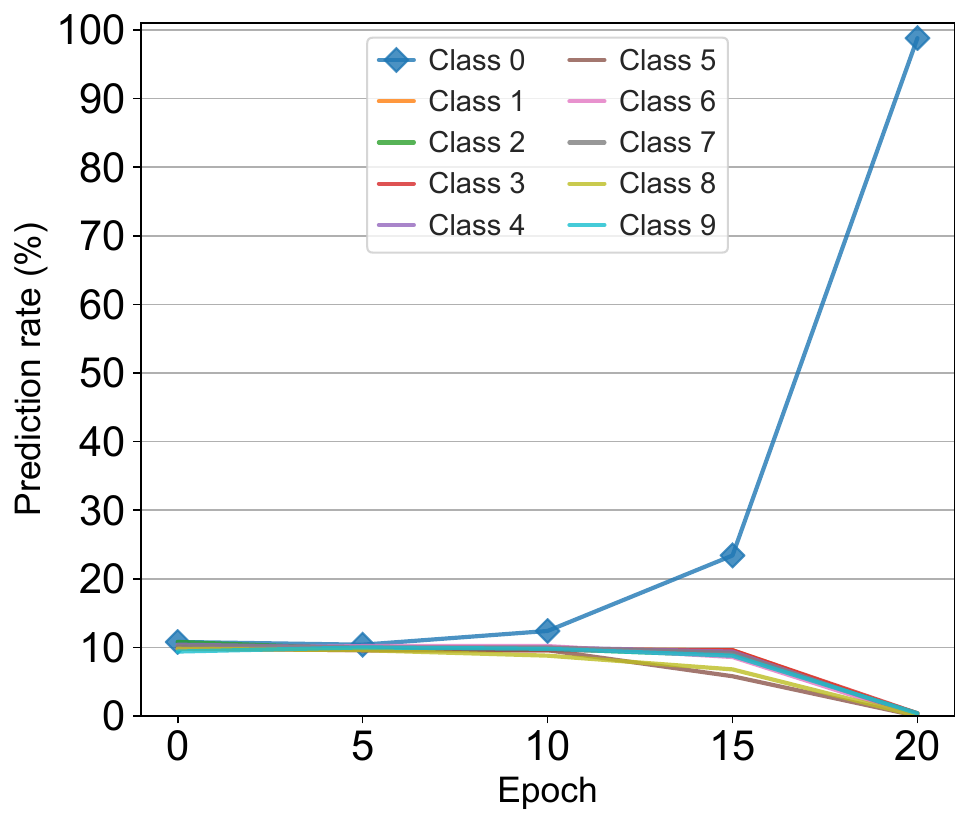}
        \includegraphics[width=0.23\linewidth]{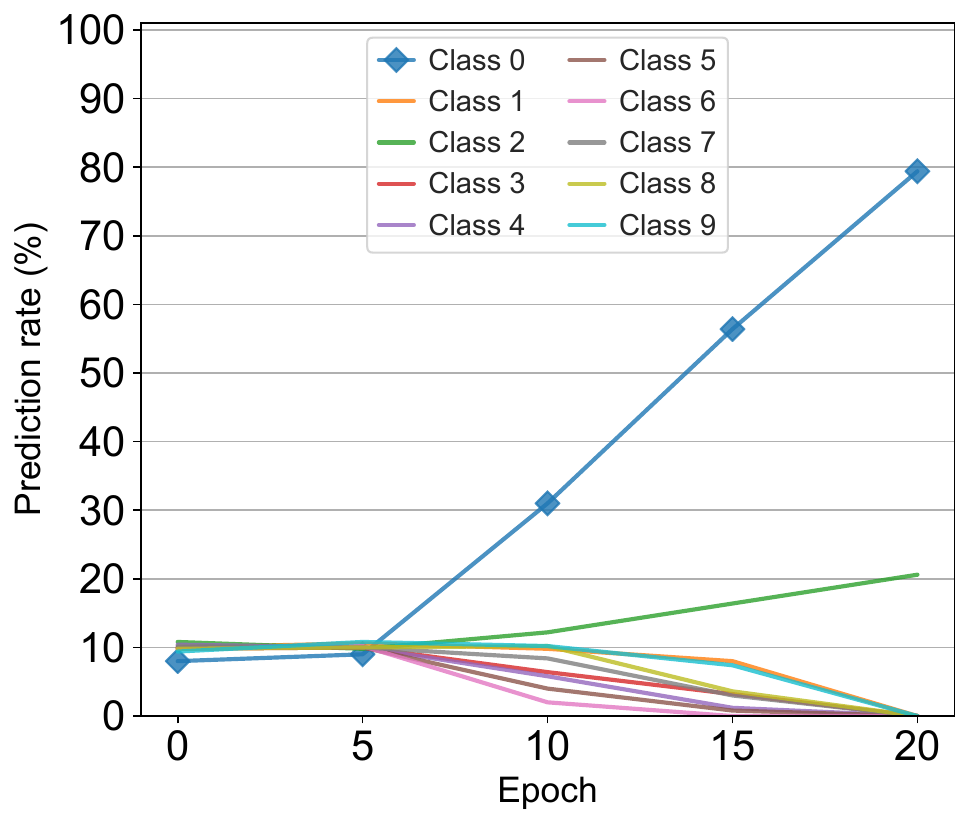}
        \includegraphics[width=0.23\linewidth]{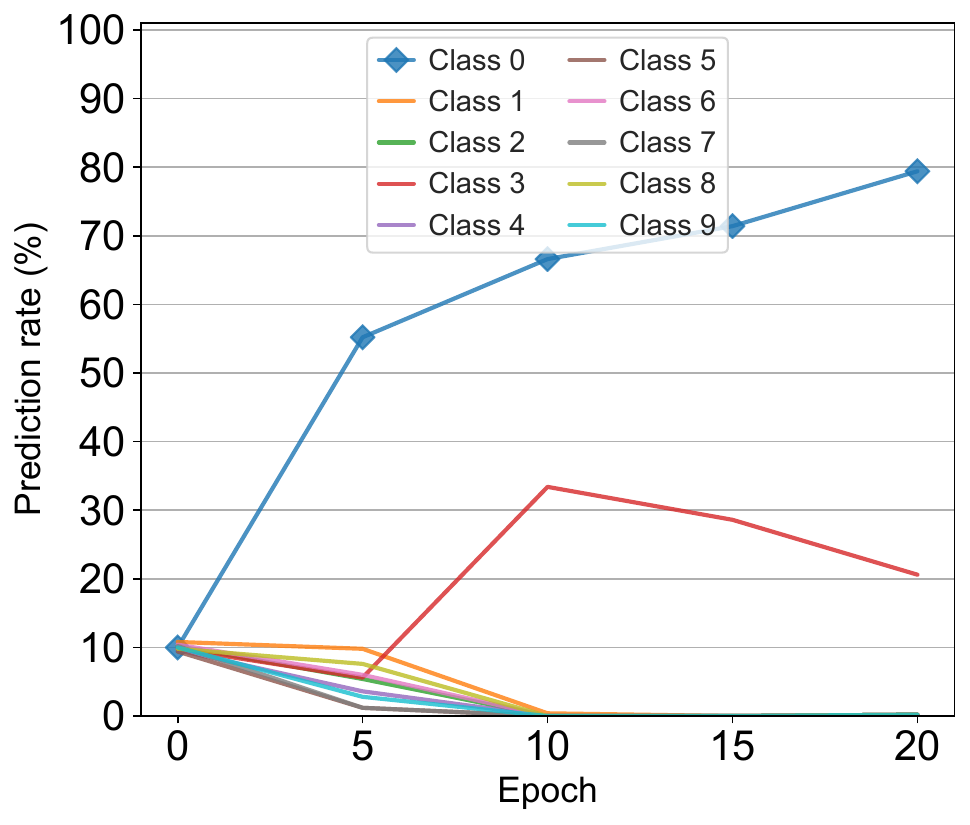}
}
\caption{Two central properties of the ``exposed model” under Clean Unlearning (CUL). The experiments were conducted with ResNet-18 and BadNets attack on the CIFAR-10 dataset.}
\label{fig:prop_pred}
\end{figure*}

\noindent\textbf{Properties of Exposed Models.} Through the backdoor exposure achieved using our CUL method, we obtain a \emph{backdoor-exposed model}. We identify two key properties of the exposed models as follows.

\begin{itemize}
\item \textbf{Property 1 (Backdoor Feature Dominance)}: The functionality of a backdoor-exposed model is dominated by the backdoor features.
\item \textbf{Property 2 (Backdoor Label Consistency)}: A backdoor-exposed model consistently predicts the backdoor target label for any input samples.
\end{itemize}

For example, for the CIFAR-10 dataset, 1\% (500) clean samples are sufficient to unlearn the clean features while exposing the backdoor neurons.  The top row in Fig. \ref{fig:prop_pred} shows that EBYD can efficiently erase the clean performance but retain the backdoor performance, as indicated by ASR and CA. Meanwhile, the bottom row of Fig. \ref{fig:prop_pred} shows that the exposure operation simultaneously exposed the backdoor label.
Notably, unlearning can be safely terminated when the performance of the model on the defense data $\mathcal{D}_{d}$ is close to a random guess. 
We defer the results of other exposure techniques to Section \ref{sec:more exposure techniques} where it shows that CUL remains the best among these techniques.

\begin{algorithm} [!tp]
	\renewcommand{\algorithmicrequire}{\textbf{Input:}}
	\renewcommand{\algorithmicensure}{\textbf{Output:}}
	\caption{Backdoor Exposure}
	\begin{algorithmic}[1]
        \REQUIRE A backdoored model $f_{\theta}(\cdot)$ with parameter $\theta$, a backdoor exposure function $\Phi: \theta \rightarrow \theta_b$, the total number of classes $K$, defense data $\mathcal{D}_{d}$, max iteration epochs $T$, clean accuracy threshold $CA_{min}$ and training loss threshold $\gamma$

        \IF{$\Phi$ is CFT}
            \FOR{$t=0$ to $T$}
            \STATE Sample a mini-batch $(\hat{\vx}_{d}, \hat{y}_{d})$ from $\hat{\gD}_{d}$
            \STATE Update $\theta_b$ by \eqref{eq:cft}
            \ENDFOR
      
        \ELSIF{$\Phi$ is CUL}
        \WHILE{Clean accuracy on $\theta_b \leq CA_{min}$ or training loss on $\theta_b \geq \gamma$}
        \STATE Sample a mini-batch $(\vx_{d}, y_{d})$ from $\gD_{d}$
        \STATE Update $\theta_b$ by \eqref{eq:cul}
        \ENDWHILE

        \ELSIF{$\Phi$ is Pruning}
        \STATE $\mathbf{m}^{\kappa} = [1]^n$  \; \# initialized mask to be all ones
        \STATE Update $\vm^{\kappa}$ and reinitialize top-n $\vm^{\kappa}$ values into zero 
        \STATE $\theta_b \leftarrow$ iterative magnitude pruning, i.e. $\theta_b = \mathbf{m}^{\kappa} \cdot \theta$

        \ELSIF{$\Phi$  is AWP}
        \STATE $\theta_b \leftarrow$ calculate perturbation $\delta$ to $\theta$ by \eqref{eq:awp}
        \ENDIF
          
      \STATE Backdoor label: $y_{t} = \underset{K}{\arg \max} \ f(\boldsymbol{x}_d,y_d;\ \theta_b)$
      \ENSURE $\theta_b$, $y_{t}$
	\end{algorithmic}
		\label{alg:exposure}
\end{algorithm}

\subsection{Other Backdoor Exposure Techniques}\label{sec:more exposure techniques}
Following \eqref{eq:EBYD_max}, here we extend our exploration from CUL to existing fine-tuning, pruning, and weight perturbation techniques. We find that these techniques can also be effective when adapted for backdoor exposure.

\subsubsection{Confusion Fine-tuning}
Previous studies have shown that backdoored models exhibit certain resilience against fine-tuning due to the inactivity of backdoor neurons when exposed to a small portion of clean defense samples \cite{li2021neural}. 
This means that with careful control, we might be able to segregate the backdoor functionality via fine-tuning.
This inspires us to propose a \textit{Confusion Fine-Tuning (CFT)} method that uncovers backdoors by fine-tuning the model on a few mislabeled clean samples. Specifically, given a deliberately mislabeled dataset $(\vx_d, \hat{y}_d) \in \hat{\gD}_d$ with modified labels $\hat{y} = \textit{Random}({1, 2, \cdots, K})$, where $K$ is the total number of classes, the optimization objective for CFT can be formulated as:
\begin{equation}\label{eq:cft}
\min_{\theta_c} \E_{(\vx_d,\hat{y}_d)\in \hat{\gD}_d} \; \|\Ls(f(\vx_d, \hat{y}_d; \theta_c\cup\theta_b)) - \gamma\|,
\end{equation} 
where $\theta$ represents the parameters of model $f$, and $\Ls$ denotes the cross-entropy loss. Following the above formulation, we will show that CFT can also erase the clean functionality while preserving the backdoor functionality. \\

\subsubsection{Model Sparsification via Pruning}
Model pruning aims to extract a sparse sub-network from the original dense network without degrading the model's performance. We denote $\mathbf{m}^{\kappa} \in \{0, 1\}^d$ as a binary mask applied to $\theta$ to indicate the locations of pruned weights (represented by zeros in $\mathbf{m}^{\kappa}$) and unpruned weights (represented by non-zeros in $\mathbf{m}^{\kappa}$). To expose the backdoor functionality, we 1) first initial $\mathbf{m}^{\kappa}$ to be all ones and then update $\mathbf{m}^{\kappa}$ on the clean subset, and then 2) iteratively prune the neurons from the model to obtain a sparse model, i.e., $\hat{\theta}_c = (\mathbf{m}^{\kappa} \odot \theta)$, which is defined as:
\begin{equation}\label{eq:pruning}
\max_{\hat{\theta}_c} \E_{(\vx_d,y_d)\in\gD_d} \; \| \Ls(f((\vx_d,y_d) ; \hat{\theta}_c \cup \theta_b)) - \gamma\|,
\end{equation} 
where the top-$n$ values in $\mathbf{m}^{\kappa}$ are initialized to be zeros and used to remove the clean neurons. Therefore, a value close to 0 in the final mask indicates the pruned clean neurons, while a value close to 1 indicates the remaining backdoor-related neurons. We observe that as the pruning rate increases, there exists a pruned model with a very high ASR and low CA. \\

\subsubsection{Adversarial Weight Perturbation (AWP)}
AWP was initially proposed for adversarial training~\cite{wu2021adversarial}. It improves the robust generalization of adversarial training by smoothing the loss landscape of the model.
Here, we adapt AWP to expose backdoor neurons from a backdoored model. We adversarially perturb the model weight parameters using AWP to maximize the model's loss on the clean defense data.
Formally, the perturbations on the model parameters can be defined as follows:
\begin{equation}\label{eq:awp}
\max_{\hat{\theta}_c} \E_{\vx_d,y_d)\in\gD} \; \| \Ls(f((\vx_d,y_d) ; \hat{\theta}_c \cup \theta_b)) - \gamma\|,
\end{equation} 
where $\hat{\theta}_c = (1+\boldsymbol{\delta}) \odot \theta_c$, $\boldsymbol{\delta}$ represents the perturbation to the model weight $\theta$, and $\Ls$ denotes the cross-entropy loss. We optimize the neuron perturbations $\delta$ to increase the loss on the clean data $(\vx_d, y_d) \in \gD_d$. Interestingly, we find that if the perturbation is well-balanced, it can effectively reduce the CA while maintaining a very high ASR on backdoor samples. The lower CA and almost unchanged ASR indicate successful backdoor exposure, as the functionality of the backdoor behavior is preserved. \\

\subsection{Measuring Backdoor Exposure}
To quantitatively assess different exposure methods, here we introduce a metric called \textit{Backdoor Exposure Metric (BEM)} to measure the effect of backdoor exposure. 
The BEM score is calculated based on the ASR and CA results of the exposed model $\theta_b$ over the first $t$ exposure epochs. Formally, BEM is defined as:
\begin{align}
BEM = \frac{\frac{1}{t} \sum_{i=0}^{t-1} \left(ASR({\theta^{i}_{b}}) - CA({\theta^{i}_{b}})\right)}{\frac{1}{t} \sum_{i=0}^{t-1} ASR({\theta^{i}_{b}})}.
\end{align}
Intuitively, BEM measures the effect of preserving ASR while erasing CA, relative to the original ASR. Note that the ASR and CA used to calculate BEM are both averaged over the first $t$ exposure epochs to obtain a more stable result. A higher BEM score indicates more effective exposure, and vice versa.

\section{Unified Defense with Exposed Model}

The above backdoor exposure step of EBYD can be viewed as an upstream task while the subsequent detection and removal tasks in the defense step are the downstream tasks. By successfully exposing the backdoor in the upstream phase, all downstream methods can target the same objective, thereby creating a comprehensive defense framework. Below, we describe how the exposed model can be utilized to enhance backdoor model detection, backdoor sample detection, and backdoor removal.

\subsection{Enhancing Backdoor Sample Detection} 
STRIP \cite{gao2019strip} observed certain differences in output entropy between benign and malicious examples and proposed to detect backdoor samples based on the prediction entropy gap. The predictions with the lower entropy imply a backdoor sample. However, advanced backdoor attacks such as those with full-image trigger patterns can violate its assumption, leading to an unclear entropy gap. To improve the identification of backdoor samples, we propose an extension of the original STRIP method by replacing the original model with the exposed model $\theta_b$ obtained through backdoor exposure. The enhanced STRIP method, based on the entropy summation of all $N$ perturbed inputs, can be formulated as:
\begin{equation} \label{equ:exp_strip}
\mathbb{H}_{sum}=-\sum_{n=1}^{n=N} \sum_{i=1}^{i=K} y_{i} \times \log _{2} f(\hat{\vx}; \ \theta_b),
\end{equation}
where $\hat{\vx}$ is the perturbed input by superimposing various image patterns and $K$ is the number of total labels. \\
 
\noindent\textbf{Remarks.} A key assumption for successful detection is that the backdoor-exposed model $\theta_b$, which contains the most information about the backdoor triggers, will predict significantly higher entropy for given inputs compared to a clean model. Consequently, a higher entropy implies a greater likelihood of an input being a backdoor sample. By extending and enhancing the original STRIP method, we improve its ability in detecting backdoor samples. An empirical analysis is presented in Section \ref{sec:EBYD_detection}.

\begin{algorithm} [!tp]
\renewcommand{\algorithmicrequire}{\textbf{Input:}}
\renewcommand{\algorithmicensure}{\textbf{Output:}}
\caption{Expose Before You Defend (EBYD)}
\label{alg:EBYD}
\begin{algorithmic}[1]
\REQUIRE Victim model $f_{\theta}(\cdot)$ with parameters $\theta$, defense dataset $\mathcal{D}_{d}$, dynamic threshold $DT$ in [0, 1]
\STATE Sample defense data $(\vx_{d}, y_{d})$ from $\mathcal{D}_{d}$
\STATE \# \textbf{Stage 1: Backdoor Exposure}
\STATE Obtain the backdoor-exposed model $f_{\theta_{b}}$ via Alg. \ref{alg:exposure}

\STATE \# \textbf{Stage 2: Backdoor Defense}

\STATE // \textit{Backdoor Sample Detection}
    \STATE Solve entropy-based detection on $f_{\theta_{b}}$ via Eq. \ref{equ:exp_strip}

\STATE // \textit{Backdoor Model Detection}
    \STATE Solve trigger-reversed optimization on $f_{\theta_{b}}$ via Eq. \ref{equ:exp_nc}

\STATE // \textit{Backdoor Removal}
\STATE \textit{\# 1) Recovering clean accuracy}
    \STATE Initialize the mask: $\mathbf{m}_r = [1]^n$ 
    \REPEAT
        \STATE $\mathbf{m}_r = \mathbf{m}_r - \eta \frac{\partial \mathcal{L}(f(X_{d}, Y_{d};\ \mathbf{m}_r \odot \theta_b))}{\partial \mathbf{m}_r}$
        \STATE $\mathbf{m}_r = clip_{[0,1]}(\mathbf{m}_r)$  \; \# 0-1 clipping
    \UNTIL{convergence}
\STATE \textit{\# 2) Pruning backdoor neurons}
    \STATE Binarize the mask: $\mathbf{m}_r \leftarrow \mathbb{I}\left(\mathbf{m}_r > DT \right)$
    \STATE Purified parameters $\hat{\theta}=\mathbf{m}_r \odot \theta$
    
\ENSURE Purified model $f_{\hat{\theta}}$

\end{algorithmic}
\end{algorithm}

\subsection{Enhancing Backdoor Model Detection}

Trigger inversion based defenses represent a prevalent detection paradigm for identifying backdoored models. Among them, one of the most well-known and foundational methods is Neural Cleanse (NC). Specifically, NC detects backdoored models by reverse engineering the trigger pattern through constrained optimization. The optimization process of NC is defined as:
\begin{equation} \label{equ:4}
    \begin{aligned}
\min _{\boldsymbol{m}, \boldsymbol{\Delta}}  \Ls (y_{t}^{k}, f(\hat{\vx};\ \theta))+\lambda \cdot|\boldsymbol{m}|,
\end{aligned}
\end{equation}
where $\hat{\vx} = (1-m) \odot \vx+m \odot \Delta$ represents the operation that applies reversed-trigger $(m, \Delta)$ into the clean input $\vx$ , $\lambda$ is the balancing parameter of the trigger size, and $k$ is the index of all target labels. 

As highlighted in previous works \cite{liu2019abs, hu2021trigger}, NC suffers from two major drawbacks: 1) It requires reverse engineering all class labels to identify the backdoor label, which can be extremely time-consuming when the total number of classes is high. 2) Due to the entanglement of clean and backdoor features, it exhibits low fidelity of reversed triggers, leading to failed detection against advanced attacks. \\

Fortunately, the two properties of the exposed model can help solve the above two drawbacks of NC, i.e., backdoor feature dominance and backdoor label consistency.
Formally, let $\theta_b$ denote the parameters of the backdoor-exposed model and $y_{t}$ is the potential trigger label, and then \eqref{equ:4} can be reformulated  as:
\begin{equation} \label{equ:exp_nc}
    \begin{aligned}
\min _{\boldsymbol{m}, \boldsymbol{\Delta}}  \ell (y_{t}, f(\hat{\vx};\ \theta_b))+\lambda \cdot|\boldsymbol{m}|. \\
\end{aligned}
\end{equation}

\noindent\textbf{Remarks.} Compared to the original NC methods, our proposed method demonstrates several advantages: (1) Efficient inference of potential backdoor target labels without any prior assumptions about the trigger type, shape, and size.
(2) Direct identification of backdoored models based on the exposed backdoor label, i.e., a backdoor label indicates the existence of a backdoor trigger. We will demonstrate how this combination can significantly enhance the performance of NC-like detection in Section \ref{sec:EBYD_detection}.

\subsection{Enhancing Backdoor Removal}
In this section, we propose a novel backdoor removal method as the last defense operation of EBYD to remove the backdoor neurons in the exposed model. The method is called \emph{Recover-Pruning (EBYD-RP)}.
Given a backdoor-exposed model, EBYD-RP first recovers the clean functionality of the model with a learnable neural mask on the clean defense data and then identifies and prunes the backdoor neurons based on the learned mask. \\

EBYD-RP first defines a neural mask for all neurons in the exposed model and then updates the mask by solving the following optimization problem:
\begin{equation} \label{equ:exp_rnp}
\begin{aligned}
\min_{\mathbf{m}_r \in[0,1]^{n}} \mathcal{L}(f(x_d ;\mathbf{m}_r \odot \theta_b)),
\end{aligned}
\end{equation}
where $\mathcal{L}$ is the cross-entropy loss, $x_d \in \mathcal{D}_{c}$ is the defense data,  $\theta_b$ are the parameters of the backdoor-exposed model obtained via CUL, and $\mathbf{m}_r$ is a mask with the same dimension as $\theta_b$ and initialized to be all ones. To allow the mask differentiable, we apply continuous relaxation to $\mathbf{m}_r$ and project it into the range of $[0, 1]^{n}$. The minimization process defined in \eqref{equ:exp_rnp} recovers the exposed model's clean performance by updating a recovery mask on the neurons. The mask helps locate neurons that change the most during the recovery process, which will be determined as backdoor neurons. \\

EBYD-RP was designed based on our observation that during the recovery process, the backdoor neurons tend to change more than the clean neurons to compensate for the clean performance loss caused by backdoor exposure (e.g., CUL). This is because the backdoor neurons are functionality-irrelevant neurons that are largely repurposed during the recovery process. Thus, by optimizing the exposed model again on the clean defense data, we can recover the accuracy of clean neurons while simultaneously disentangling and pruning a certain percentage of the backdoor neurons. \\

Based on the learned mask  $\mathbf{m}_r$, the optimal pruning rate can be flexibly determined via dynamic thresholding in $[0, 1]$. The idea is to prune as many neurons as possible until the drop in the clean accuracy becomes unacceptable. After optimization, we prune the neurons by setting the mask value which smaller than the threshold to be 0. A high value close to 1 in $\mathbf{m}_r$ indicates that the neuron is indeed important for clean performance, while a low value close to 0 means that the neuron is indeed a backdoor neuron. And neurons with smaller values in $\mathbf{m}_r$ should be pruned in this case. Note that, the best pruning rate can be pre-specified or flexibly determined via dynamic thresholds \cite{wu2021adversarial}. In our experiments, we adopt dynamic thresholding as our default setting, unless otherwise stated. \\

\noindent\textbf{Overall Algorithm.} Algorithm \ref{alg:EBYD} outlines the two defense steps of EBYD: backdoor exposure and backdoor defense. In the \textit{backdoor exposure} step, the framework reveals hidden backdoor functionalities in the victim model, $f_{\theta}(\cdot)$. Using specialized techniques, it produces a backdoor-exposed model, $f_{\theta_{b}}$, which uncovers malicious features embedded in the original model. This step is essential for the subsequent defense tasks. The \textit{backdoor defense} step performs three key tasks: backdoor sample detection, backdoor model detection, and backdoor removal. Particularly, backdoor sample/model detection aims to identify and filter out harmful inputs or backdoored models. If a backdoored model is identified, our EBYD framework will work to remove the backdoor using EBYD-RP, which involves iterative optimization and pruning to recover the model's clean accuracy and eliminate backdoor triggers. The final outcome is a purified model. Overall, our EBYD framework is highly versatile, suitable for various backdoor defense scenarios, and offers a holistic, modular solution to enhance the safety of AI systems.

\begin{table}[!tp]
\small
\centering
\caption{Detailed information of the datasets and classifiers used in our experiments.}
\label{tab:dataset_model}
\begin{adjustbox}{width=\linewidth}
\begin{tabular}{ccccc}
\toprule
Dataset & Labels & Type & Training Sizes & Classifier \\  \midrule
CIFAR-10 & 10 & Image & 50000 & ResNet-18  \\
ImageNet subset & 20 & Image & 26000 & ResNet-50 \\
SST-2 & 2 & Text & 6920 & Bert-base-uncased  \\
IMDB & 2 & Text & 22500 & Bert-base-uncased \\
Twitter & 2 & Text & 69633 & Bert-base-uncased  \\
AG's News & 4 & Text & 11106 & Bert-base-uncased \\
\bottomrule
\end{tabular}
\end{adjustbox}
\end{table}

\begin{figure*}[!tp]
\centering
\includegraphics[width=0.92\linewidth]{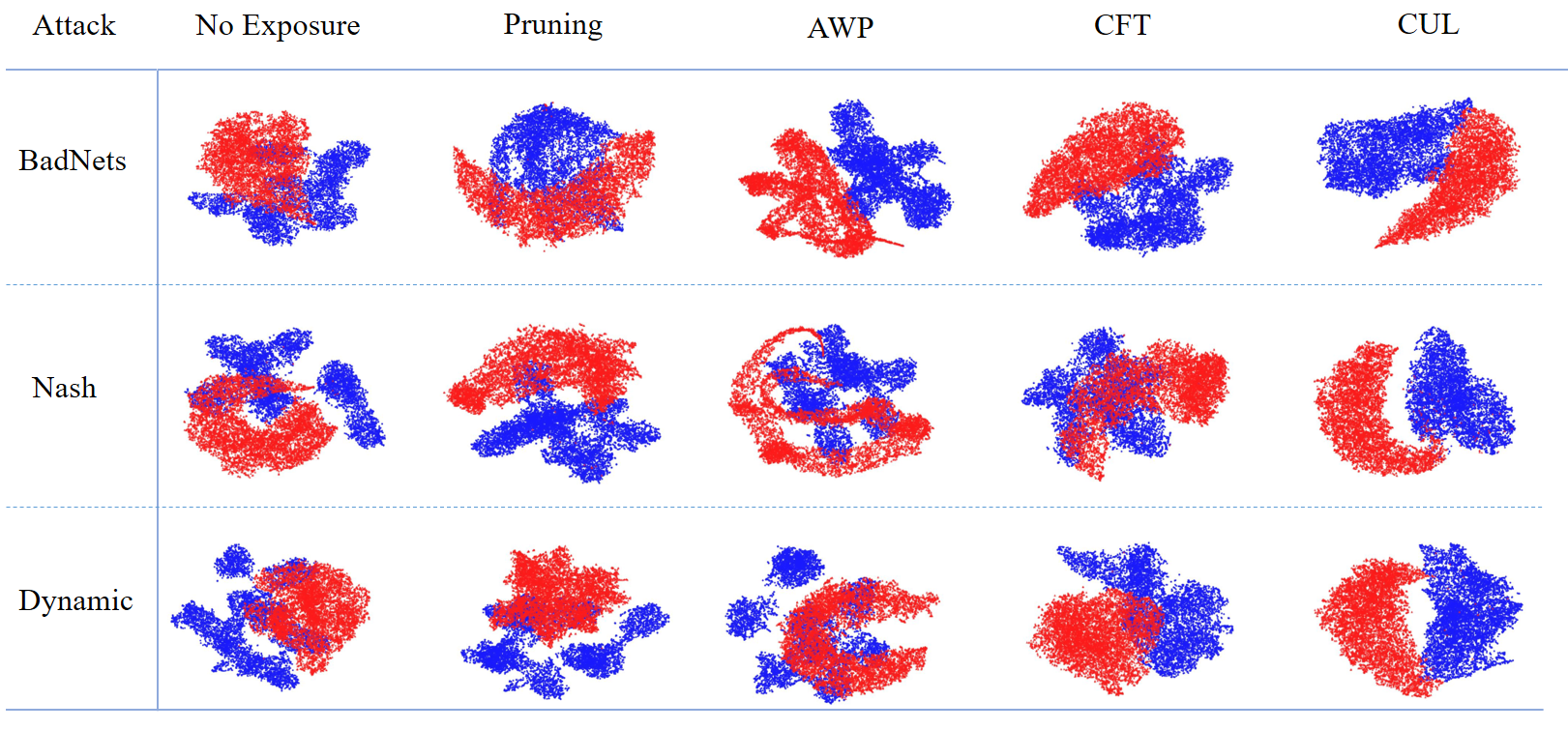}
\caption{A t-SNE visualization of the decoupled clean (blue) and backdoor (red) features by 4 backdoor exposure techniques.}
\label{fig:exp_tsne}
\end{figure*}

\begin{table*}[!tp]
\centering
\caption{The exposure index of 4 backdoor exposure techniques measured by BEM. The best average results are \textbf{boldfaced}.}
\begin{adjustbox}{width=0.9\linewidth}
\begin{tabular}{c|cccccccccc|c}
\toprule
\textbf{Exposure Index} & \textbf{OnePixel} & \textbf{BadNets} & \textbf{Trojan} & \textbf{Blend} & \textbf{SIG} & \textbf{CL} & \textbf{Smooth} & \textbf{Nash} & \textbf{Dynamic} & \textbf{WaNet} & \textbf{Average} \\ \hline
Pruning & 0.66 & 0.68 & 0.63 & 0.62 & 0.67 & 0.60 & 0.63 & 0.62 & 0.68 & 0.64 & 0.64 \\
AWP & 0.97 & 0.99 & 0.87 & 0.72 & 0.89 & 0.61 & 0.81 & 1.00 & 0.98 & 1.00 & 0.88 \\
CFT & 0.73 & 0.87 & 0.69 & 0.71 & 0.72 & 0.87 & 0.79 & 0.70 & 0.68 & 0.85 & 0.76 \\
CUL & 0.88 & 0.91 & 0.99 & 0.86 & 0.91 & 0.87 & 0.88 & 0.99 & 0.99 & 0.95 & \textbf{0.92} \\ \bottomrule
\end{tabular}
\end{adjustbox}
\label{tab:exposure_index}
\end{table*}

\section{Experiments}\label{sec:experiments}
\subsection{Experimental Setup}
\noindent\textbf{Datasets and Models.}
Our experiments consider both image and text classification tasks. For image classification, we consider two commonly used datasets CIFAR-10 \cite{krizhevsky2009learning} and ImageNet \cite{deng2009imagenet} subset (the first 20 classes), with the ResNet \cite{he2016deep} model. 
For text classification, we consider four classical NLP datasets including SST-2 \cite{socher2013recursive}, IMBD \cite{maas2011learning}, Twitter \cite{founta2018large}, and AG’s News \cite{zhang2015character}, with the transformer model BERT \cite{devlin2018bert}. The experimental details can be found in the appendix. \\

\noindent\textbf{Attack Setup.} We evaluate our defense against both image and text backdoor attacks. For image attacks, we chose 10 representative backdoor attacks on image classification: OnePixel \cite{tran2018spectral}, BadNets \cite{gu2017badnets}, Trojan \cite{liu2018trojaning}, Blend \cite{chen2017targeted}, SIG \cite{barni2019new}, Adv \cite{turner2019clean}, Smooth \cite{zeng2021adversarial}, Nash \cite{liu2019abs}, Dynamic \cite{nguyen2020input}, and WaNet \cite{nguyen2021wanet}. Fig. \ref{fig:backdoor_example} shows a few examples of backdoor triggers used in our experiments.  For text attacks, we consider 6 textual backdoor attacks on text classification: BadNet-RW \cite{gu2017badnets}, BadNet-SL \cite{chen2021badnl}, Syntactic \cite{qi2021hidden}, SOS \cite{yang2021rethinking},
RIPPLE \cite{kurita2020weight}, and LWP \cite{li2021layerwise}.
We used the official implementations of these attacks and followed their suggested settings in the original papers, including trigger pattern, trigger size, and backdoor label. Table \ref{tab:attacks_overview} summarizes the detailed settings of these attacks. \\

\noindent\textbf{EBYD Setup.} 
In EBYD, we explore and evaluate four backdoor exposure techniques, including \textit{CFT}, \textit{Pruning}, \textit{AWP}, and our proposed \textit{CUL}. On the defense side, we demonstrate how the backdoor-exposed model can be adopted to enhance the defense performance for three representative backdoor defense methods: Neural Cleanse (NC) \cite{wang2019neural}, STRIP \cite{gao2019strip}, and our proposed Recover-Pruning (RP). This covers the entire spectrum of defense scenarios involving backdoor model detection, backdoor sample detection, and backdoor model removal. All defenses have limited access to only 500 clean samples held out from the CIFAR-10 training set (or ImageNet subset using the same data augmentation techniques, i.e., random crop ($\text{padding} = 4$) and horizontal flipping, as discussed in the attack settings. The detailed defense setup is described in the appendix. \\

\noindent\textbf{Performance Metrics.}
We adopt three metrics to evaluate the defense methods: 1) Detection Rate (DR), which represents the success rate of the defense in identifying the backdoor label or backdoored model. More specifically, we use the Area under the ROC curve (AUROC) as the detection metric; 2) Clean Accuracy (CA), which measures the model's accuracy on clean test data; and 3) Attack Success Rate (ASR), which reflects the model's accuracy on backdoored test data.
Note that we have removed the samples whose ground-truth labels are the same as the backdoor label, ensuring that a perfect defense will achieve a nearly zero ASR while maintaining a high CA.

\begin{figure*}[!tp]
	\centering
    \subfigure[OnePixel]{\includegraphics[width=.18\textwidth]{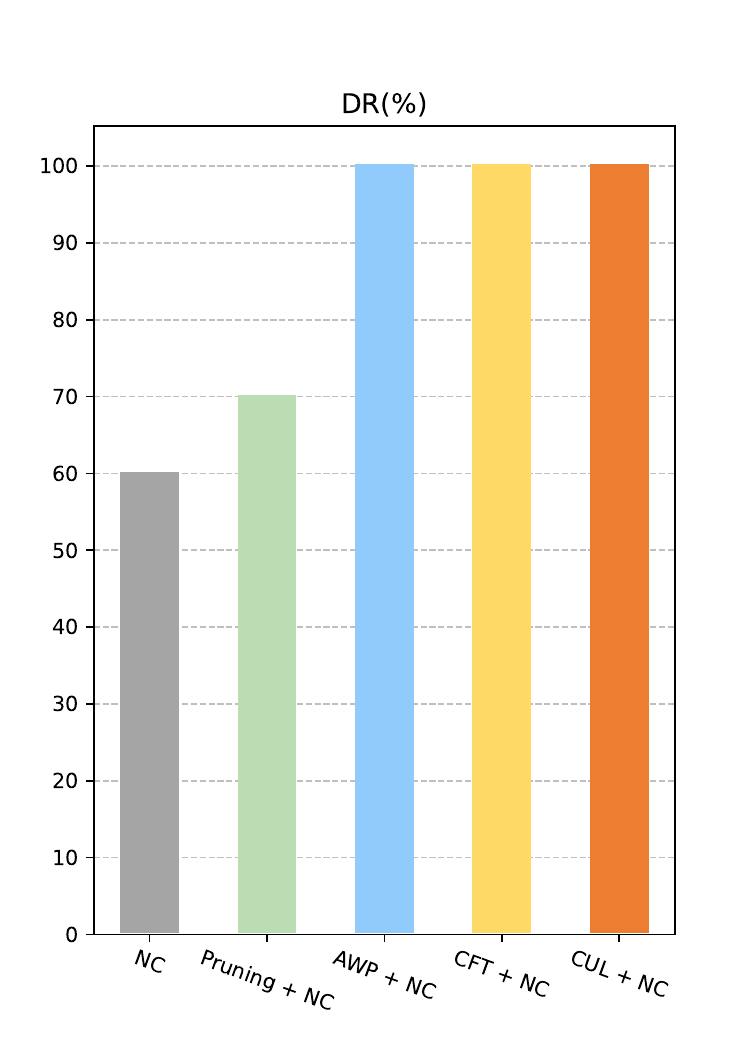}}
    \subfigure[BadNets]{\includegraphics[width=.18\textwidth]{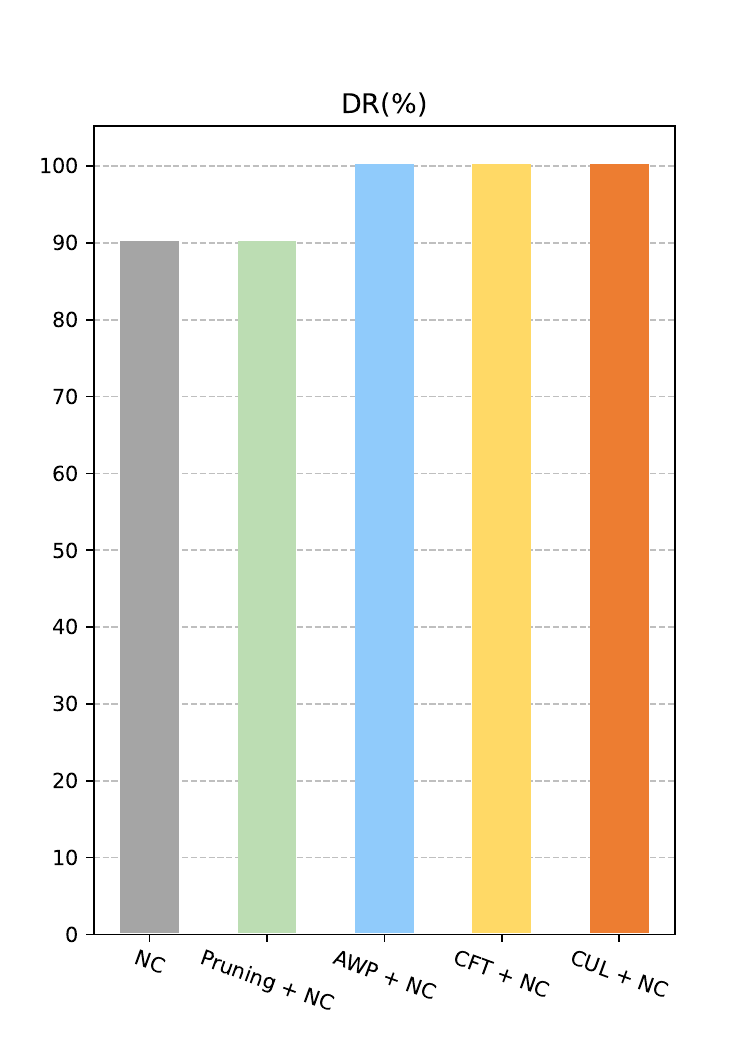}}
    \subfigure[Trojan]{\includegraphics[width=.18\textwidth]{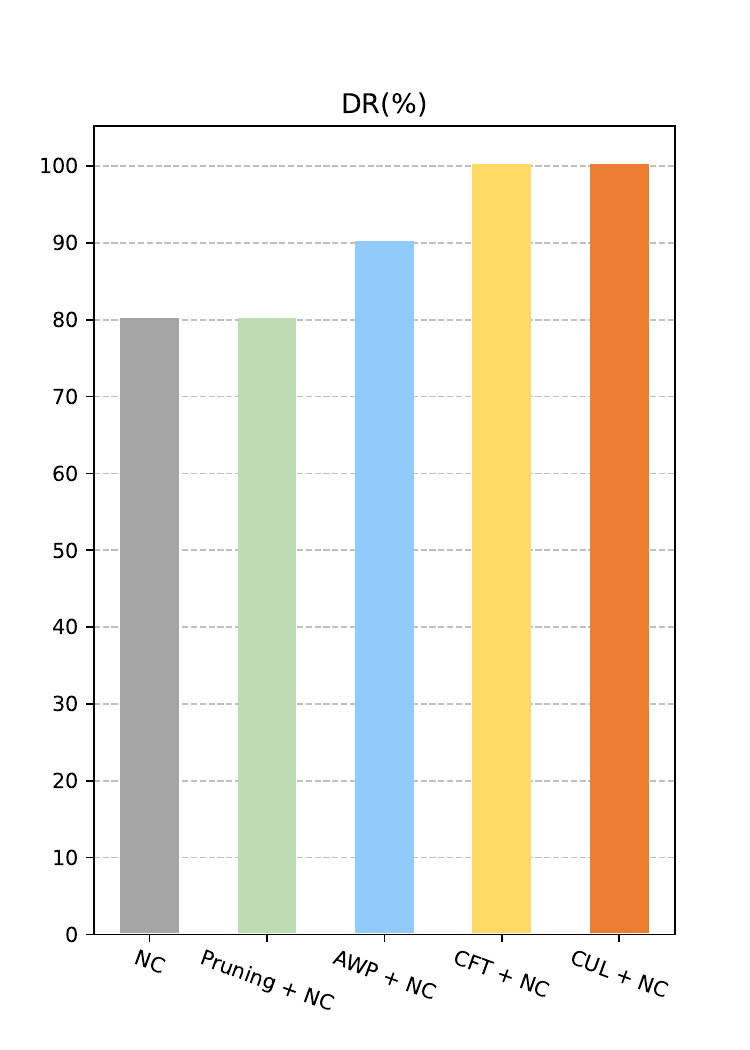}}
    \subfigure[Blend]{\includegraphics[width=.18\textwidth]{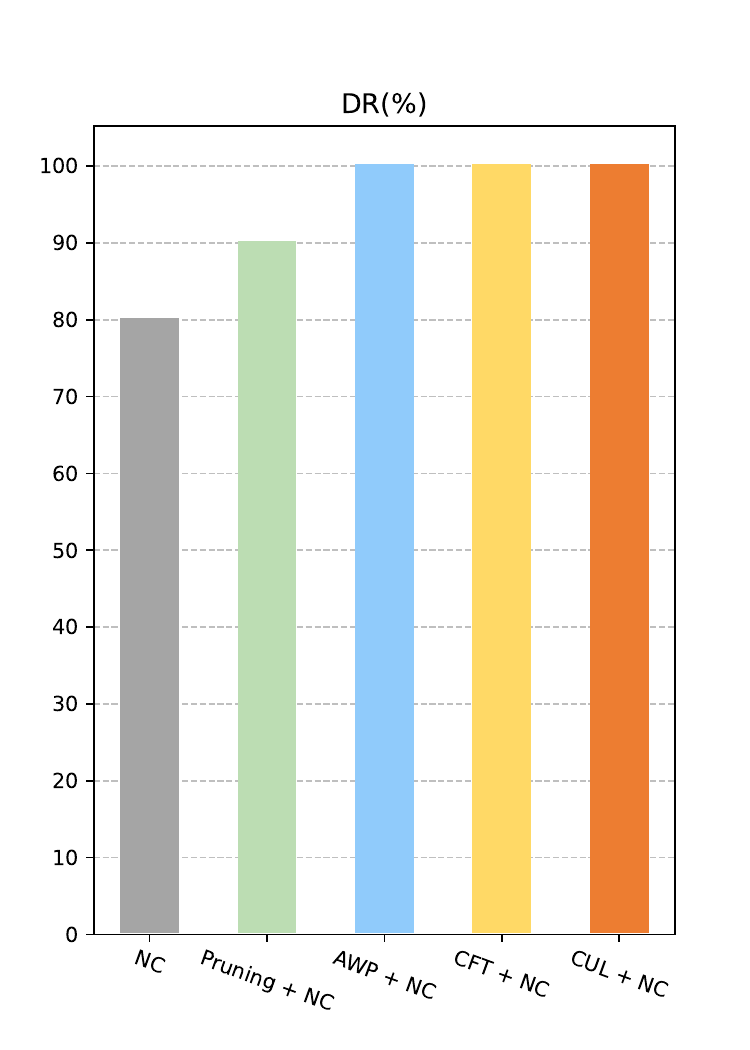}}
   \subfigure[SIG]{\includegraphics[width=.18\textwidth]{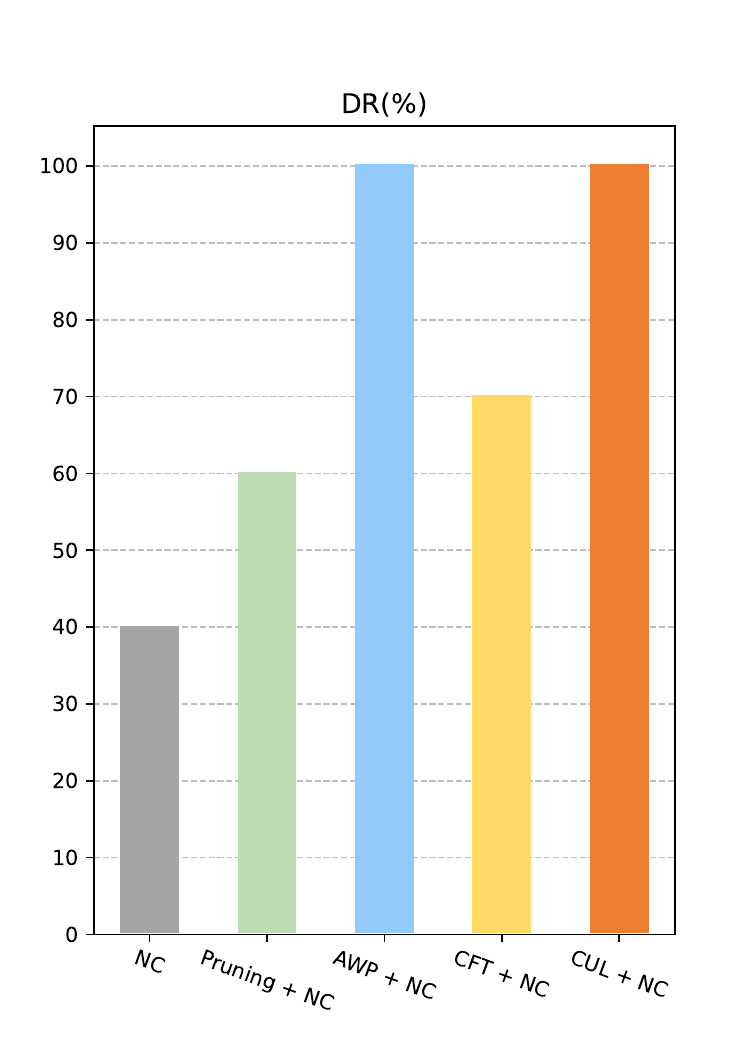}}
    \subfigure[Adv]{\includegraphics[width=.18\textwidth]{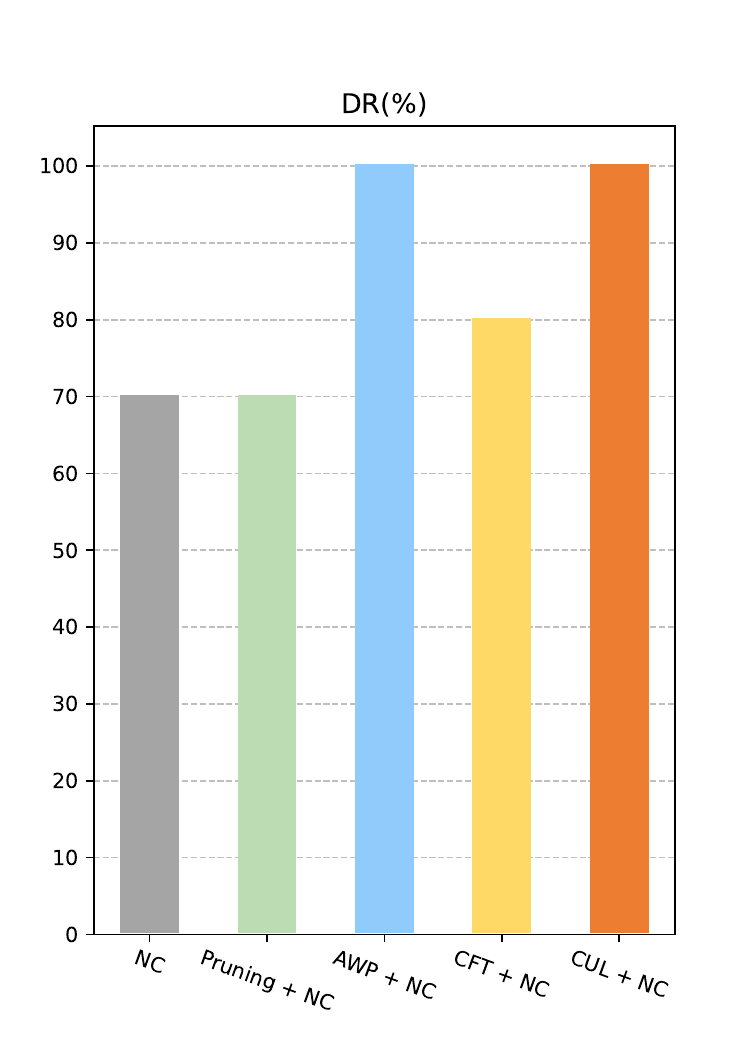}}
    \subfigure[Smooth]{\includegraphics[width=.18\textwidth]{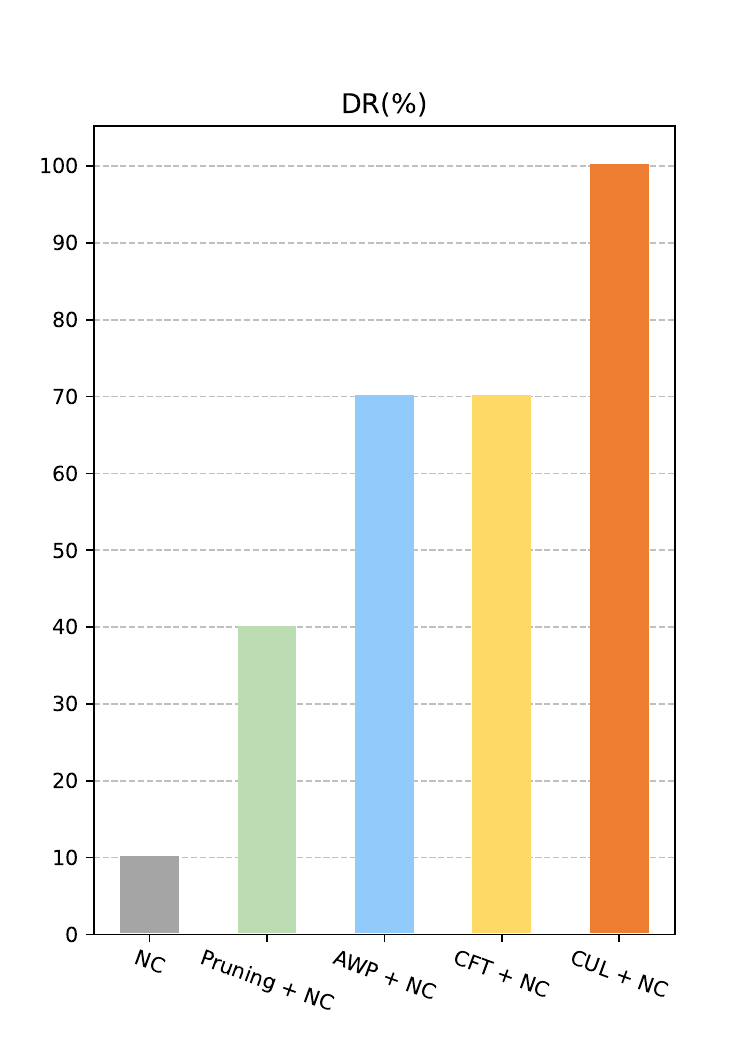}}
    \subfigure[Nash]{\includegraphics[width=.18\textwidth]{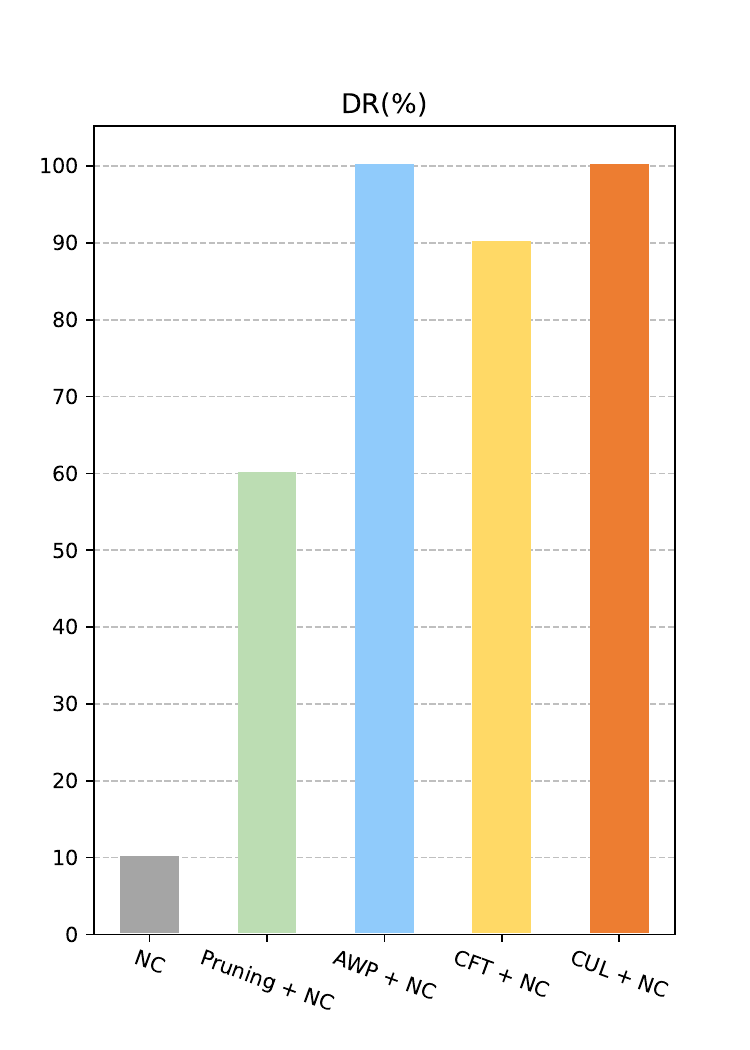}}
    \subfigure[Dynamic]{\includegraphics[width=.18\textwidth]{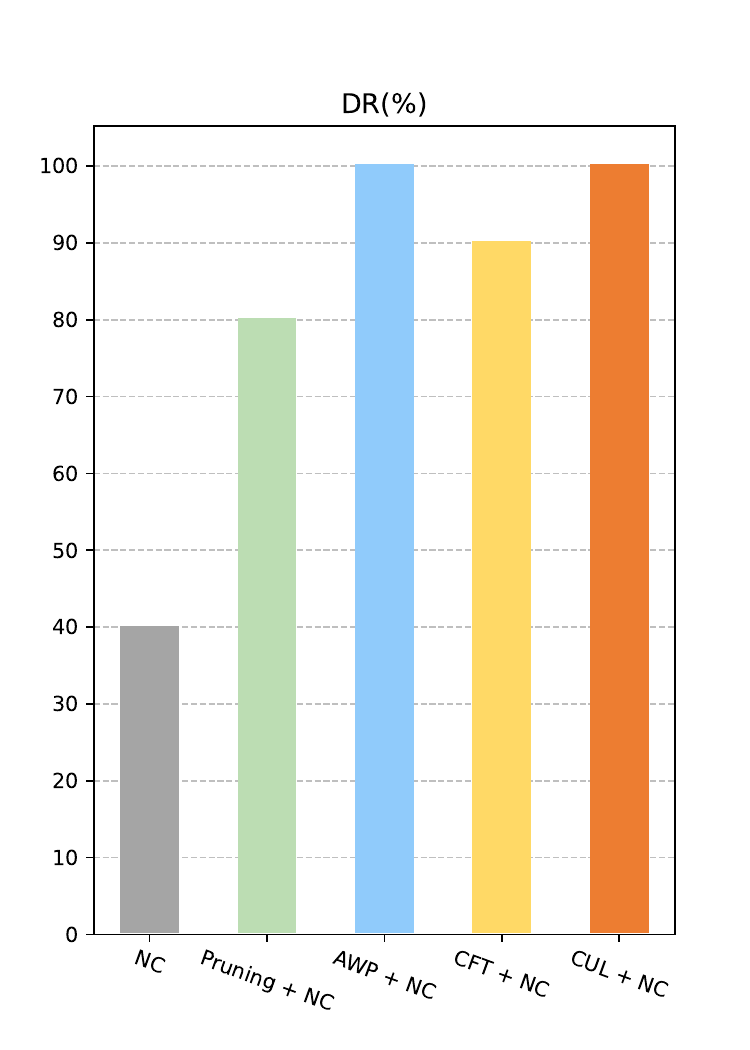}}
    \subfigure[WaNet]{\includegraphics[width=.18\textwidth]{imgs/NC_plt/SIG.pdf}}
	\caption{ The detection performance of `X+NC' against 10 backdoor attacks on CIFAR-10.  DR (\%): AUROC rate. }
	\label{fig:EBYD_NC}
\end{figure*}

\begin{table*}[!tp]        
\centering
\caption{The detection performance of `X+STRIP' against 10 backdoor attacks on CIFAR-10.  DR (\%): AUROC rate. The best average results are \textbf{boldfaced}.}
\begin{adjustbox}{width=0.9\linewidth}
\begin{tabular}{c|cccccccccc|c}
\hline
\textbf{DR (\%)} & \textbf{OnePixel} & \textbf{BadNets} & \textbf{Trojan} & \textbf{Blend} & \textbf{SIG} & \textbf{Adv} & \textbf{Smooth} & \textbf{Nash} & \textbf{Dynamic} & \textbf{WaNet} & \textbf{Average} \\ \hline
STRIP & 48.31 & 66.47 & 47.04 & 33.72 & 28.65 & 97.82 & 47.42 & 40.94 & 74.69 & 54.02 & 53.91 \\ \hline
Pruning+STRIP & 63.12 & 87.06 & 80.13 & 69.14 & 50.34 & 68.84 & 64.23 & 61.46 & 86.95 & 73.66 & \textbf{70.49} \\
AWP+STRIP & 91.07 & 95.67 & 89.99 & 89.99 & 79.88 & 99.99 & 98.88 & 81.8 & 89.99 & 78.62 & \textbf{89.59} \\
CFT+STRIP & 81.12 & 99.17 & 98.81 & 79.94 & 89.87 & 91.21 & 94.54 & 76.14 & 87.96 & 80.42 & \textbf{87.92} \\
CUL+STRIP & 91.57 & 98.16 & 95.12 & 89.98 & 81.29 & 92.93 & 94.13 & 85.59 & 97.24 & 83.36 & \textbf{90.94} \\ \hline
\end{tabular}
\end{adjustbox}
\label{tab:EBYD_STRIP}

\end{table*}

\subsection{Evaluating and Understanding Backdoor Exposure}
We show the effective backdoor exposure metrics (BEM) for four different backdoor exposure strategies: Pruning, AWP, CFT, and CUL. This experiment was conducted on three backdoored ResNet-18 models subjected to attacks including BadNets, Nash, and Dynamic on the CIFAR-10 dataset with backdoor label 0. Note that the defense data $\mathcal{D}_{d}$ contains only 500 clean samples. The results in Table \ref{tab:exposure_index} show that among the four backdoor exposure techniques, our proposed CUL performs the best, achieving the highest average BEM score of 0.92. In comparison, other techniques—Pruning, AWP, and CFT—attain lower average exposure indices of 0.64, 0.88, and 0.76, respectively. The CUL method, which unlearns the model's clean functionality on a few clean samples, effectively isolates backdoor features while minimally affecting the backdoor functionality. In contrast, other methods may disrupt or inadequately decouple these features. For instance, AWP adds perturbations to the model parameters, which can interfere with backdoor features, whereas fine-tuning or pruning techniques damage the backdoor functionality. Overall, the BEM results underscore the superiority of our CUL method in backdoor exposure, highlighting its effectiveness in preserving the integrity of backdoor features while facilitating their exposure for defensive purposes. In Fig. \ref{fig:prop_pred}, we use CUL as an example and showcase the exposure performance for downstream tasks in terms of CA and ASR.

We plot the decoupled clean-backdoor feature distributions by different exposure techniques in Fig. \ref{fig:exp_tsne} using t-SNE \cite{tsne}. This leads to several key insights: 
1) For simpler attacks like BadNets, all exposure techniques successfully decouple and reveal backdoor features. Specifically, the separation between backdoor features and clean features increases, indicating the effective exposure of backdoor features within the model.
2) Against more advanced attacks, such as Nash and Dynamic attacks, the effectiveness of different exposure techniques varies. For instance, Nash exhibits a more intricate feature distribution, with significant overlap between clean and backdoor-related features, making it challenging for techniques like CFT and AWP to isolate backdoor features. In contrast, pruning-based techniques show moderate success, increasing the distance between clean and backdoor features but still exhibiting some entanglement, as a small fraction of backdoor features remain within the clean feature space.
3) Notably, CUL demonstrates to be a more stable and efficient backdoor decoupling method, outperforming other techniques by effectively isolating backdoor features across all attack types.

\begin{table*}[tp!]        
\renewcommand{\arraystretch}{1.4}
\renewcommand\tabcolsep{4.5pt}
\centering
\caption{The performance of our `EBYD' against 10 backdoor attacks under different defense stages including backdoor exposure via EBYD, clean recovery, and EBYD-aided Recover-Pruning (EBYD-RP). The best average results are \textbf{boldfaced}.}
\begin{adjustbox}{width=0.95\linewidth}
\begin{tabular}{c|c|cc|cc|cc|cc|cc|cc|cc|cc|cc|cc|cc}
\hline
\multirow{2}{*}{\textbf{Defense Stage}} & \multirow{2}{*}{\textbf{EBYD Strategy}} & \multicolumn{2}{c|}{\textbf{OnePixel}} & \multicolumn{2}{c|}{\textbf{BadNets}} & \multicolumn{2}{c|}{\textbf{Trojan}} & \multicolumn{2}{c|}{\textbf{Blend}} & \multicolumn{2}{c|}{\textbf{SIG}} & \multicolumn{2}{c|}{\textbf{Adv}} & \multicolumn{2}{c|}{\textbf{Smooth}} & \multicolumn{2}{c|}{\textbf{Nash}} & \multicolumn{2}{c|}{\textbf{Dynamic}} & \multicolumn{2}{c|}{\textbf{WaNet}} & \multicolumn{2}{c}{\textbf{Average}} \\ \cline{3-24} 
 &  & \textbf{ASR} & \textbf{CA} & \textbf{ASR} & \textbf{CA} & \textbf{ASR} & \textbf{CA} & \textbf{ASR} & \textbf{CA} & \textbf{ASR} & \textbf{CA} & \textbf{ASR} & \textbf{CA} & \textbf{ASR} & \textbf{CA} & \textbf{ASR} & \textbf{CA} & \textbf{ASR} & \textbf{CA} & \textbf{ASR} & \textbf{CA} & \textbf{ASR} & \textbf{CA} \\ \hline
\multirow{5}{*}{FT} & NO & 14.03 & 91.58 & 22.13 & 92.06 & 13.80 & 92.49 & 43.81 & 92.29 & 24.59 & 92.34 & 100.00 & 91.94 & 93.03 & 92.27 & 30.67 & 92.67 & 30.91 & 92.64 & 9.90 & 91.69 & 38.29 & 92.20 \\ \cline{2-24} 
 & Pruning & 16.76 & 87.57 & 19.89 & 82.30 & 14.83 & 89.42 & 31.72 & 89.00 & 16.87 & 85.42 & 100.00 & 86.47 & 58.18 & 87.30 & 25.89 & 89.46 & 18.92 & 83.61 & 12.20 & 87.93 & 31.53 & 86.85 \\
 & AWP & 9.34 & 88.19 & 10.14 & 87.16 & 8.71 & 87.20 & 14.39 & 88.90 & 12.39 & 86.16 & 39.99 & 87.03 & 34.37 & 86.49 & 20.33 & 86.62 & 17.71 & 88.83 & 4.26 & 85.41 & 17.16 & 87.20 \\
 & CFT & 5.29 & 90.86 & 28.43 & 91.43 & 19.46 & 92.52 & 8.80 & 91.11 & 6.08 & 91.72 & 100.00 & 91.30 & 88.72 & 91.26 & 56.82 & 92.28 & 15.08 & 92.38 & 5.96 & 90.90 & 33.46 & \textbf{91.58} \\
 & CUL & 1.39 & 90.28 & 1.72 & 88.22 & 3.36 & 91.94 & 11.14 & 89.58 & 0.24 & 90.86 & 6.38 & 81.88 & 23.22 & 91.48 & 3.60 & 90.87 & 15.38 & 91.63 & 2.40 & 90.19 & \textbf{6.88} & 89.69 \\ \hline
\multirow{5}{*}{ANP} & NO & 7.79 & 91.82 & 14.96 & 90.63 & 6.06 & 92.49 & 1.99 & 92.66 & 1.27 & 92.38 & 6.56 & 91.46 & 7.19 & 91.44 & 13.28 & 92.57 & 29.37 & 92.61 & 7.59 & 91.12 & 9.61 & 91.92 \\ \cline{2-24} 
 & Pruning & 1.22 & 91.79 & 6.42 & 91.20 & 3.66 & 92.27 & 0.71 & 91.40 & 1.26 & 91.27 & 3.19 & 91.24 & 2.70 & 87.36 & 3.39 & 92.62 & 14.47 & 91.33 & 3.80 & 91.58 & 4.08 & 91.21 \\
 & AWP & 1.44 & 91.81 & 2.26 & 90.80 & 1.68 & 92.02 & 0.77 & 92.40 & 1.18 & 92.02 & 5.28 & 91.41 & 2.23 & 91.39 & 2.80 & 92.63 & 14.79 & 92.32 & 6.59 & 90.68 & 3.90 & \textbf{91.75} \\
 & CFT & 3.88 & 91.99 & 2.62 & 91.11 & 1.56 & 92.20 & 0.56 & 91.31 & 0.52 & 91.40 & 1.83 & 91.72 & 6.33 & 89.32 & 4.79 & 92.78 & 10.91 & 90.32 & 2.98 & 91.54 & 3.60 & 91.37 \\
 & CUL & 1.13 & 91.82 & 1.67 & 91.63 & 1.37 & 92.83 & 5.76 & 91.27 & 1.04 & 89.17 & 2.63 & 87.77 & 0.43 & 90.72 & 2.84 & 91.59 & 8.31 & 91.38 & 2.12 & 90.90 & \textbf{2.73} & 90.91 \\ \hline
\multirow{5}{*}{ABL} & NO & 99.62 & 81.37 & 3.04 & 86.11 & 3.81 & 87.46 & 16.23 & 84.06 & 10.09 & 88.27 & 99.99 & 87.00 & 99.94 & 85.09 & 8.80 & 88.47 & 5.84 & 88.63 & 28.37 & 83.93 & 37.57 & 86.04 \\ \cline{2-24} 
 & Pruning & 40.21 & 88.96 & 2.53 & 87.27 & 3.10 & 82.11 & 13.25 & 85.27 & 9.12 & 86.55 & 100.00 & 87.46 & 99.01 & 87.69 & 4.32 & 89.23 & 3.38 & 86.51 & 10.89 & 82.23 & 28.58 & 86.33 \\
 & AWP & 0.20 & 91.61 & 0.04 & 91.08 & 0.76 & 92.39 & 2.63 & 83.10 & 8.73 & 87.29 & 0.88 & 91.27 & 5.01 & 90.78 & 0.43 & 91.06 & 2.76 & 91.79 & 0.88 & 90.36 & \textbf{2.23} & 90.07 \\
 & CFT & 0.20 & 91.61 & 1.03 & 91.54 & 2.69 & 92.23 & 1.82 & 81.60 & 6.48 & 75.83 & 0.99 & 91.58 & 7.48 & 90.67 & 5.82 & 82.99 & 4.07 & 92.06 & 3.26 & 88.92 & 3.38 & 87.90 \\
 & CUL & 0.22 & 91.62 & 0.06 & 91.47 & 0.04 & 92.04 & 1.87 & 88.74 & 2.30 & 89.47 & 94.57 & 91.07 & 3.14 & 89.97 & 1.58 & 91.26 & 2.79 & 92.18 & 0.27 & 90.30 & 10.68 & \textbf{90.81} \\ \hline
\multirow{5}{*}{RP} & NO & 4.40 & 91.76 & 30.70 & 90.08 & 1.94 & 92.21 & 1.98 & 92.20 & 1.90 & 91.54 & 7.56 & 86.63 & 1.31 & 90.50 & 5.77 & 92.42 & 16.22 & 91.63 & 3.51 & 91.44 & 7.53 & 91.04 \\ \cline{2-24} 
 & Pruning & 3.08 & 91.78 & 1.48 & 90.16 & 1.24 & 91.61 & 0.03 & 92.21 & 0.41 & 90.89 & 7.41 & 89.44 & 1.24 & 90.61 & 3.89 & 92.50 & 10.01 & 92.17 & 2.56 & 91.21 & 3.14 & 91.26 \\
 & AWP & 1.31 & 91.24 & 1.31 & 91.01 & 1.00 & 90.79 & 0.07 & 92.29 & 0.09 & 90.16 & 4.84 & 91.28 & 2.53 & 88.50 & 2.43 & 91.27 & 7.39 & 91.67 & 3.62 & 90.01 & 2.46 & 90.82 \\
 & CFT & 1.52 & 91.72 & 1.82 & 90.37 & 1.06 & 91.97 & 0.10 & 92.34 & 0.31 & 90.66 & 1.16 & 91.28 & 3.52 & 90.74 & 4.92 & 92.32 & 10.72 & 91.72 & 2.82 & 91.03 & 2.80 & 91.32 \\
 & CUL & 1.07 & 91.67 & 0.44 & 90.37 & 0.99 & 92.41 & 0.01 & 92.24 & 0.09 & 90.12 & 0.58 & 91.07 & 1.60 & 91.54 & 1.39 & 91.06 & 5.49 & 92.71 & 2.04 & 90.80 & \textbf{1.37} & \textbf{91.40} \\ \hline
\end{tabular}
\end{adjustbox}
 \label{tab:EBYD_Removal}

\end{table*}

\subsection{Enhancing Backdoor Detection with Exposed Model} \label{sec:EBYD_detection}

Backdoor detection involves both model-level and sample-level detection, and thus, we address both aspects in our evaluation. We consider the representative model-level detection approach, Neural Cleanse (NC), and the sample-level approach, STRIP, as examples to demonstrate how the backdoor-exposed model contributes to their detection performance. For simplicity, we denote `X+NC' and `X+STRIP' as the original NC and STRIP methods applied to the exposed model by one of the exposure techniques, respectively. For instance, `Pruning+NC' refers to applying NC detection on the exposing model through the pruning technique.  \\

\noindent\textbf{Backdoor Model Detection.} Fig. \ref{fig:EBYD_NC} illustrates the detection performances of `X+NC' against 10 backdoor attacks on CIFAR-10. It is evident that, in most cases, `X+NC' achieves a significant improvement in the average detection rate (DR) compared to the original NC. In general, `CUL+NC' achieves the best results, improving the average DR by more than 20\% across all 10 attacks, which is significantly better than other combinations such as `CFT+NC' and `AWP+NC'. The reason behind this is that the better exposure of the backdoor features (illustrated in Fig. \ref{fig:exp_tsne}) makes backdoor identification easier and more precise. \\

We find that each exposure technique has its own limitations against certain attacks. For instance, even though `Pruning+NC' and `CFT+NC' have the best overall performance against simple attacks like BadNets, Trojan, and Blend, they are weaker in defending against more stealthy and invisible attacks such as SIG, Nash, Dynamic, and WaNet, with a low DR ranging from $40\%$ to 70\%. 
This is likely due to an insufficient exposure of the backdoor features, leading to misaligned backdoor trigger recovery, as shown in Fig. \ref{fig:exp_tsne}. 
For the Smooth attack, `AWP+NC' shows much poorer performance than `CUL+NC', with a 30\% performance drop. 
We speculate that adversarial perturbation on model weights cannot effectively disentangle the backdoor features when perturbation-based backdoor triggers closely match the clean inputs. 
Finally, `Pruning+NC' exhibits the poorest overall performance, with an average DR of less than 70\% against most attacks, indicating that pruning-based exposure is ineffective against backdoor attacks.

In summary, `X+NC', especially `CUL+NC', achieved a superior performance against all backdoor attacks compared to the original NC. We emphasize that exposing backdoor features within a backdoored model holds promise for more precise detection. Examples of the recovered triggers with the exposed models can be found in Fig. \ref{fig:EBYD_trigger} in the appendix. \\

\begin{table*}[tp!]        
\renewcommand{\arraystretch}{1.4}
\renewcommand\tabcolsep{4.5pt}
\centering
\caption{The performance of our EBYD against 10 backdoor attacks in different backdoor defense tasks including backdoor exposure (BE), backdoor model detection (BMD), backdoor sample detection (BSD), and backdoor removal (BR). The best average results are \textbf{boldfaced}.}
\begin{adjustbox}{width=0.95\linewidth}
\begin{tabular}{c|cc|cc|cc|cc|cc|cc|cc|cc|cc|cc|cc}
\hline
\textbf{Method} & \multicolumn{2}{c|}{\textbf{OnePixel}} & \multicolumn{2}{c|}{\textbf{BadNets}} & \multicolumn{2}{c|}{\textbf{Trojan}} & \multicolumn{2}{c|}{\textbf{Blend}} & \multicolumn{2}{c|}{\textbf{SIG}} & \multicolumn{2}{c|}{\textbf{Adv}} & \multicolumn{2}{c|}{\textbf{Smooth}} & \multicolumn{2}{c|}{\textbf{Nash}} & \multicolumn{2}{c|}{\textbf{Dynamic}} & \multicolumn{2}{c|}{\textbf{WaNet}} & \multicolumn{2}{c}{\textbf{Average}} \\ \hline
\rowcolor[HTML]{C0C0C0} 
\textbf{BE} & \multicolumn{2}{c|}{\cellcolor[HTML]{C0C0C0}Index} & \multicolumn{2}{c|}{\cellcolor[HTML]{C0C0C0}Index} & \multicolumn{2}{c|}{\cellcolor[HTML]{C0C0C0}Index} & \multicolumn{2}{c|}{\cellcolor[HTML]{C0C0C0}Index} & \multicolumn{2}{c|}{\cellcolor[HTML]{C0C0C0}Index} & \multicolumn{2}{c|}{\cellcolor[HTML]{C0C0C0}Index} & \multicolumn{2}{c|}{\cellcolor[HTML]{C0C0C0}Index} & \multicolumn{2}{c|}{\cellcolor[HTML]{C0C0C0}Index} & \multicolumn{2}{c|}{\cellcolor[HTML]{C0C0C0}Index} & \multicolumn{2}{c|}{\cellcolor[HTML]{C0C0C0}Index} & \multicolumn{2}{c}{\cellcolor[HTML]{C0C0C0}Index} \\ \hline
None & \multicolumn{2}{c|}{0.07} & \multicolumn{2}{c|}{0.09} & \multicolumn{2}{c|}{0.08} & \multicolumn{2}{c|}{0.08} & \multicolumn{2}{c|}{0.08} & \multicolumn{2}{c|}{0.09} & \multicolumn{2}{c|}{0.07} & \multicolumn{2}{c|}{0.08} & \multicolumn{2}{c|}{0.00} & \multicolumn{2}{c|}{0.08} & \multicolumn{2}{c}{0.07} \\
CUL & \multicolumn{2}{c|}{0.88} & \multicolumn{2}{c|}{0.91} & \multicolumn{2}{c|}{0.99} & \multicolumn{2}{c|}{0.86} & \multicolumn{2}{c|}{0.91} & \multicolumn{2}{c|}{0.87} & \multicolumn{2}{c|}{0.86} & \multicolumn{2}{c|}{0.99} & \multicolumn{2}{c|}{0.99} & \multicolumn{2}{c|}{0.95} & \multicolumn{2}{c}{\textbf{0.92}} \\ \hline
\rowcolor[HTML]{C0C0C0} 
\textbf{BMD} & \multicolumn{2}{c|}{\cellcolor[HTML]{C0C0C0}DR} & \multicolumn{2}{c|}{\cellcolor[HTML]{C0C0C0}DR} & \multicolumn{2}{c|}{\cellcolor[HTML]{C0C0C0}DR} & \multicolumn{2}{c|}{\cellcolor[HTML]{C0C0C0}DR} & \multicolumn{2}{c|}{\cellcolor[HTML]{C0C0C0}DR} & \multicolumn{2}{c|}{\cellcolor[HTML]{C0C0C0}DR} & \multicolumn{2}{c|}{\cellcolor[HTML]{C0C0C0}DR} & \multicolumn{2}{c|}{\cellcolor[HTML]{C0C0C0}DR} & \multicolumn{2}{c|}{\cellcolor[HTML]{C0C0C0}DR} & \multicolumn{2}{c|}{\cellcolor[HTML]{C0C0C0}DR} & \multicolumn{2}{c}{\cellcolor[HTML]{C0C0C0}DR} \\ \hline
NC & \multicolumn{2}{c|}{60} & \multicolumn{2}{c|}{90} & \multicolumn{2}{c|}{80} & \multicolumn{2}{c|}{80} & \multicolumn{2}{c|}{40} & \multicolumn{2}{c|}{70} & \multicolumn{2}{c|}{10} & \multicolumn{2}{c|}{10} & \multicolumn{2}{c|}{40} & \multicolumn{2}{c|}{40} & \multicolumn{2}{c}{68} \\
CUL+NC & \multicolumn{2}{c|}{100} & \multicolumn{2}{c|}{100} & \multicolumn{2}{c|}{100} & \multicolumn{2}{c|}{100} & \multicolumn{2}{c|}{100} & \multicolumn{2}{c|}{100} & \multicolumn{2}{c|}{100} & \multicolumn{2}{c|}{100} & \multicolumn{2}{c|}{100} & \multicolumn{2}{c|}{100} & \multicolumn{2}{c}{\textbf{100}} \\ \hline
\rowcolor[HTML]{C0C0C0} 
\textbf{BSD} & \multicolumn{2}{c|}{\cellcolor[HTML]{C0C0C0}AUROC} & \multicolumn{2}{c|}{\cellcolor[HTML]{C0C0C0}AUROC} & \multicolumn{2}{c|}{\cellcolor[HTML]{C0C0C0}AUROC} & \multicolumn{2}{c|}{\cellcolor[HTML]{C0C0C0}AUROC} & \multicolumn{2}{c|}{\cellcolor[HTML]{C0C0C0}AUROC} & \multicolumn{2}{c|}{\cellcolor[HTML]{C0C0C0}AUROC} & \multicolumn{2}{c|}{\cellcolor[HTML]{C0C0C0}AUROC} & \multicolumn{2}{c|}{\cellcolor[HTML]{C0C0C0}AUROC} & \multicolumn{2}{c|}{\cellcolor[HTML]{C0C0C0}AUROC} & \multicolumn{2}{c|}{\cellcolor[HTML]{C0C0C0}AUROC} & \multicolumn{2}{c}{\cellcolor[HTML]{C0C0C0}AUROC} \\ \hline
STRIP & \multicolumn{2}{c|}{48.31} & \multicolumn{2}{c|}{66.47} & \multicolumn{2}{c|}{47.04} & \multicolumn{2}{c|}{33.72} & \multicolumn{2}{c|}{28.65} & \multicolumn{2}{c|}{97.82} & \multicolumn{2}{c|}{47.42} & \multicolumn{2}{c|}{40.94} & \multicolumn{2}{c|}{74.69} & \multicolumn{2}{c|}{54.02} & \multicolumn{2}{c}{53.91} \\
CUL+STRIP & \multicolumn{2}{c|}{91.57} & \multicolumn{2}{c|}{98.16} & \multicolumn{2}{c|}{95.12} & \multicolumn{2}{c|}{89.98} & \multicolumn{2}{c|}{81.29} & \multicolumn{2}{c|}{92.93} & \multicolumn{2}{c|}{94.13} & \multicolumn{2}{c|}{85.59} & \multicolumn{2}{c|}{97.24} & \multicolumn{2}{c|}{83.36} & \multicolumn{2}{c}{\textbf{90.94}} \\ \hline
\rowcolor[HTML]{C0C0C0} 
\textbf{BR} & ASR & CA & ASR & CA & ASR & CA & ASR & CA & ASR & CA & ASR & CA & ASR & CA & ASR & CA & ASR & CA & ASR & CA & ASR & CA \\ \hline
FT & 14.03 & 91.58 & 22.13 & 92.06 & 13.80 & 92.49 & 43.81 & 92.29 & 24.59 & 92.34 & 100.00 & 91.94 & 93.03 & 92.27 & 30.67 & 92.67 & 30.91 & 92.64 & 9.90 & 91.69 & 38.29 & 92.20 \\
ANP & 7.79 & 91.82 & 14.96 & 90.63 & 6.06 & 92.49 & 1.99 & 92.66 & 1.27 & 92.38 & 6.56 & 91.46 & 7.19 & 91.44 & 13.28 & 92.57 & 29.37 & 92.61 & 7.59 & 91.12 & 9.61 & 91.92 \\
ABL & 99.62 & 81.37 & 3.04 & 86.11 & 3.81 & 87.46 & 16.23 & 84.06 & 10.09 & 88.27 & 99.99 & 87.00 & 99.94 & 85.09 & 8.80 & 88.47 & 5.84 & 88.63 & 28.37 & 83.93 & 37.57 & 86.04 \\
RP & 4.40 & 91.76 & 30.70 & 90.08 & 1.94 & 92.21 & 1.98 & 92.20 & 1.90 & 91.54 & 7.56 & 86.63 & 1.31 & 90.50 & 5.77 & 92.42 & 16.22 & 91.63 & 3.51 & 91.44 & 7.53 & 91.04 \\
CUL+FT & 1.39 & 90.28 & 1.72 & 88.22 & 3.36 & 91.94 & 11.14 & 89.58 & 0.24 & 90.86 & 6.38 & 81.88 & 23.22 & 91.48 & 3.60 & 90.87 & 15.38 & 91.63 & 2.40 & 90.19 & \textbf{6.88} & \textbf{89.69} \\
CUL+ANP & 1.13 & 91.82 & 1.67 & 91.63 & 1.37 & 92.83 & 5.76 & 91.27 & 1.04 & 89.17 & 2.63 & 87.77 & 0.43 & 90.72 & 2.84 & 91.59 & 8.31 & 91.38 & 2.12 & 90.90 & \textbf{2.73} & \textbf{90.91} \\
CUL+ABL & 0.22 & 91.62 & 0.06 & 91.47 & 0.04 & 92.04 & 1.87 & 88.74 & 2.30 & 89.47 & 94.57 & 91.07 & 3.14 & 89.97 & 1.58 & 91.26 & 2.79 & 92.18 & 0.27 & 90.30 & \textbf{10.68} & \textbf{90.81} \\
CUL+RP & 1.07 & 91.67 & 0.44 & 90.37 & 0.99 & 92.41 & 0.01 & 92.24 & 0.09 & 90.12 & 0.58 & 91.07 & 1.60 & 91.54 & 1.39 & 91.06 & 5.49 & 92.71 & 2.04 & 90.80 & \textbf{1.37} & \textbf{91.40} \\ \hline
\end{tabular}
\end{adjustbox}
 \label{tab:EBYD_defense}
\end{table*}

\noindent\textbf{Backdoor Sample Detection.} 
STRIP identifies potential backdoor samples based on the prediction entropy between clean and backdoored outputs. In this subsection, we demonstrate how the proposed EBYD framework can significantly enhance the performance of the original STRIP defense against a wide range of stronger attacks. To adapt our EBYD for STRIP, we simply replace the original model with the backdoor-exposed model $f_{\theta_b}$ (denoted as `X+STRIP'). \\

Table \ref{tab:EBYD_STRIP} displays the average AUROC detection results against 10 backdoor attacks. Notably, all four combinations—`AWP+STRIP', `CFT+STRIP', `CUL+STRIP', and `Pruning+STRIP'—achieved an excellent average detection performance of 70.49\%, 89.59\%, 87.92\%, and 90.94\%, respectively. The AUROC for each combination outperforms the original `STRIP' by 16.58\%, 35.68\%, 34.01\%, and 37.03\%, respectively. 
This result verifies that EBYD can amplify the effectiveness of the original STRIP.
The superimposing technique used by STRIP results in high prediction entropy for clean samples and low entropy for backdoor samples. When clean functionality is removed from the model while backdoor functionality is preserved with ‘EBYD’, the difference in entropy becomes even more pronounced.
Among the four exposure techniques, `CUL+STRIP' achieves the best AUROC of 90.04\%, surpassing `AWP+STRIP' and `CFT+STRIP'. 
This is not surprising, as `CUL' is the most effective method for separating clean and backdoor features. 
Unfortunately, `Pruning+STRIP' performs poorly in terms of average AUROC, despite still being better than the original STRIP. We speculate that when neurons are removed through pruning, applying superimposing to clean samples results in more stable predictions (low entropy), causing them to exhibit characteristics similar to backdoor samples. This reduces the effectiveness of detection compared to using other ‘EBYD’ methods.
When examining specific attacks, we find that SIG, Nash, and WaNet are much more challenging to detect due to their invisible backdoor triggers. However, our EBYD brings significant detection improvements across all the combinations.

\subsection{Enhancing Backdoor Removal with Exposed Model} \label{sec:EBYD_removal}
Similar to previous experiments, backdoor removal methods can also be directly applied to exposed models. 
To fully explore the benefit of our EBYD, we test the application of backdoor removal in each defense step of EBYD, including backdoor exposure, model recovery, and EBYD-aided Recover-Pruning (EBYD-RP). The experiments were conducted on CIFAR-10 and the ImageNet-20 subset with ResNet-18/50 models. \\

Table \ref{tab:EBYD_defense} reports the defense performance against 10 backdoor attacks on CIFAR-10. 
Our proposed RP, equipped with four exposure techniques (Pruning, AWP, CFT, and CUL), achieves outstanding results, reducing the ASR of most attacks from 100\% to nearly 0\%, while incurring an average CA drop of less than 2\%. We hypothesize that the ability to expose the backdoor functionality within a model leads to more accurate localization of the backdoor neurons and thus more precise backdoor pruning. 
Notably, the intermediate recovery step (a process of standard fine-tuning) of RP (which consists of two steps: recovery and pruning) alone also demonstrates a strong defense effect, albeit slightly less effective than the full RP process. 
The recovery step alone can effectively reduce the ASR for most attacks while maintaining a high CA.
We believe this is because, during the recovery step, the backdoor neurons are fine-tuned to remedy the loss of clean accuracy caused by the unlearning. 
This observation is consistent with the findings in previous work \cite{li2023reconstructive}, which reveals that fine-tuning the exposed model alone can be an effective defense. The underlying mechanism behind this phenomenon deserves further investigation.

\subsection{Defense Performance of EBYD Against Image Attacks} \label{sec:EBYD_removal_exp}

Table \ref{tab:EBYD_defense} presents the effectiveness of our EBYD defense against 10 backdoor attacks across 4 defense tasks: backdoor exposure (BE), backdoor model detection (BMD), backdoor sample detection (BSD), and backdoor removal (BR). 
The experiments were conducted with ResNet-18  models on CIFAR-10. Overall, EBYD demonstrates the strongest defense capabilities against diverse backdoor attacks, particularly in reducing the ASR while maintaining high CA. \\

In the backdoor exposure (BE) task, the CUL technique of our EBYD significantly improves the exposure index across various attacks, such as increasing it from 0.0700 to 0.88 under the OnePixel attack, effectively revealing hidden backdoors. For backdoor model detection (BMD), the `CUL+NC' achieves a $100\%$ detection rate against all attacks, demonstrating its robustness. For backdoor sample detection (BSD), `CUL + STRIP' achieves a near-perfect performance, for example, improving the AUROC from 47.04\% to 99.99\% against the Trojan attack. For backdoor removal (BR), `CUL+RP' reduces the ASR by a considerable amount to 1.3\% on average while maintaining a high CA. \\

The consistent performance of EBYD across all these scenarios underscores its versatility and effectiveness in mitigating backdoor threats. By integrating techniques for backdoor exposure, detection, and removal, EBYD enhances model security against a variety of attack types. These results emphasize EBYD's potential for practical applications in secure AI systems, providing a comprehensive and adaptable approach to backdoor defense.

\begin{table}[!tp]
\small
\centering
\caption{The performance of our `EBYD-RP' defense against 6 textual backdoor attacks across 4 text datasets.}
\begin{adjustbox}{width=\linewidth}
\begin{tabular}{c|c|cccccc}
\hline
\multirow{2}{*}{\textbf{Datasets}} & \multirow{2}{*}{\textbf{Backdoor Attack}} & \multicolumn{2}{c}{\textbf{ONION}} & \multicolumn{2}{c}{\textbf{FT}} & \multicolumn{2}{c}{\textbf{EBYD-RP}} \\
 &  & \textbf{CA} & \textbf{ASR} & \textbf{CA} & \textbf{ASR} & \textbf{CA} & \textbf{ASR} \\ \hline
\multirow{7}{*}{SST-2} & BadNet-RW & 89.73 & 23.76 & 85.89 & 37.62 & 89.35 & 13.97 \\
 & BadNet-SL & 89.24 & 81.19 & 86.06 & 29.04 & 89.51 & 23.21 \\
 & Syntactic & 87.15 & 90.10 & 87.10 & 28.27 & 88.58 & 29.70 \\
 & SOS & 88.52 & 14.85 & 87.20 & 20.79 & 89.13 & 18.15 \\
 & RIPPLE & 88.96 & 22.00 & 86.71 & 43.67 & 88.25 & 18.04 \\
 & LWP & 88.30 & 16.17 & 87.15 & 99.34 & 87.20 & 41.47 \\ \hline
 & Average & 88.65 & 41.35 & 86.69 & 43.12 & 88.67 & \textbf{24.09} \\ \hline
\multirow{7}{*}{IMDB} & BadNet-RW & 91.75 & 14.47 & 88.31 & 20.54 & 90.80 & 14.61 \\
 & BadNet-SL & 89.81 & 90.22 & 85.79 & 21.82 & 91.72 & 23.61 \\
 & Syntactic & 91.37 & 93.21 & 88.27 & 18.03 & 90.29 & 18.96 \\
 & SOS & 90.14 & 13.54 & 88.34 & 16.39 & 92.74 & 12.61 \\
 & RIPPLE & 89.35 & 15.31 & 88.71 & 15.62 & 91.45 & 23.54 \\
 & LWP & 89.23 & 14.01 & 85.99 & 46.34 & 88.62 & 38.24 \\ \hline
 & Average & 90.28 & 40.13 & 87.57 & 23.12 & 90.94 & \textbf{21.93} \\ \hline
\multirow{7}{*}{Twitter} & BadNet-RW & 87.65 & 49.28 & 91.72 & 18.67 & 93.49 & 10.01 \\
 & BadNet-SL & 87.91 & 93.15 & 90.94 & 31.75 & 93.69 & 11.73 \\
 & Syntactic & 87.59 & 95.76 & 85.44 & 75.22 & 92.98 & 42.98 \\
 & SOS & 88.03 & 39.61 & 91.79 & 15.87 & 93.45 & 7.65 \\
 & RIPPLE & 88.00 & 53.67 & 91.51 & 19.04 & 93.59 & 9.36 \\
 & LWP & 87.38 & 54.90 & 92.51 & 41.97 & 93.84 & 20.14 \\ \hline
 & Average & 87.76 & 64.40 & 90.65 & 33.75 & 93.51 & \textbf{16.98} \\ \hline
\multirow{7}{*}{AG's News} & BadNet-RW & 93.16 & 22.42 & 90.05 & 23.65 & 91.54 & 17.37 \\
 & BadNet-SL & 92.87 & 93.95 & 89.99 & 20.81 & 90.11 & 12.30 \\
 & Syntactic & 92.96 & 96.29 & 88.70 & 26.11 & 91.46 & 15.47 \\
 & SOS & 93.36 & 14.56 & 90.25 & 11.95 & 92.04 & 11.58 \\
 & RIPPLE & 93.12 & 23.23 & 90.21 & 10.32 & 90.97 & 16.11 \\
 & LWP & 92.79 & 22.58 & 88.91 & 44.79 & 90.39 & 39.18 \\ \hline
 & Average & 93.04 & 45.51 & 89.69 & 22.94 & 91.09 & \textbf{18.67} \\ \hline
\end{tabular}
\end{adjustbox}
\label{tab:EBYD-RP_nlp}
\end{table}

\subsection{Defense Performance of EBYD Against Text Attacks} \label{sec:EBYD_nlp}
In this section, we evaluate the generalization performance of our EBYD-RP defense method in addressing textual backdoor attacks within the text classification task. We conduct experiments across 4 text datasets: SST-2, IMDB, Twitter, and AG's News, utilizing the BERT-base-uncased model. Our EBYD-RP is assessed against 6 representative textual backdoor attacks: BadNet-RW, BadNet-SL, Syntactic, SOS, RIPPLE, and LWP. For our EBYD-RP defense, we maintain a consistent setup of `CUL' as the default backdoor exposure technique. Detailed implementation information regarding the datasets and attack methods can be found in the appendix. \\

\begin{figure}[!tp]
\centering
\includegraphics[width=0.95\linewidth]{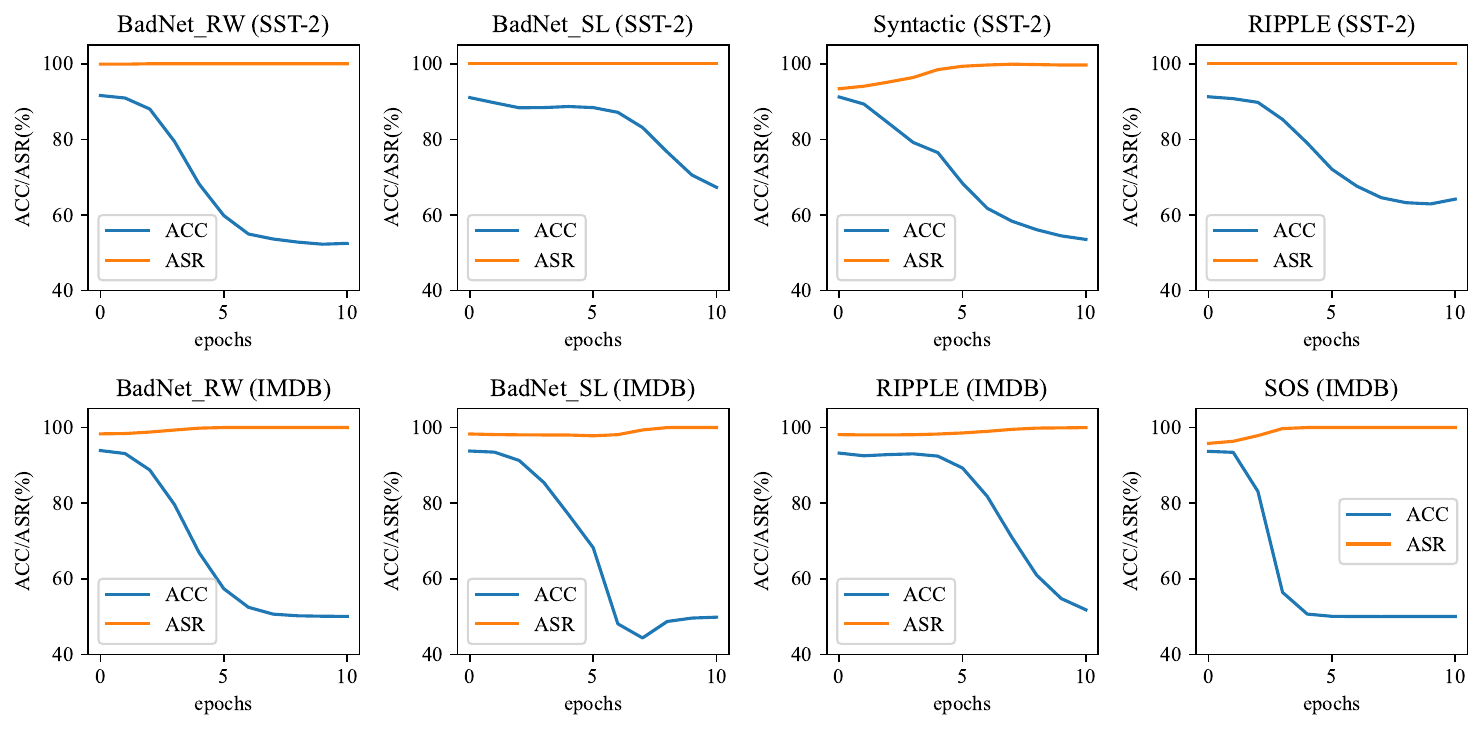}
\caption{An illustrative example of our `CUL' backdoor exposure technique against 4 textual backdoored models.}
\label{fig:EBYD_nlp}
\end{figure}

We first validate the effectiveness of EBYD in exposing the backdoor functionality within text classification models. As shown in Fig. \ref{fig:EBYD_nlp}, EBYD significantly reduces the classification accuracy of all four text-based backdoors—BadNet-RW, BadNet-SL, SOS, and RIPPLE—on the SST-2 and IMDB datasets, while maintaining the ASR (i.e., the backdoor functionality) largely unchanged. This finding indicates that the backdoor functionality can also be exposed in language models, further confirming the strong generalization capability of our EBYD framework.
Moreover, our method is highly efficient; it requires only a few epochs of unlearning (loss maximization) on a limited number of clean samples to effectively expose backdoor features.

Table \ref{tab:EBYD-RP_nlp} presents the removal results for our EBYD-RP against 6 textual backdoor attacks across 3 different datasets. Compared to existing state-of-the-art defenses like ONION, EBYD-RP achieves significant improvements in average ASR reduction: from 41.35\% to 24.09\% on SST-2, 40.13\% to 21.93\% on IMDB, 64.40\% to 16.98\% on Twitter, and 45.51\% to 18.67\%—all with a minimal drop in CA. This improvement likely results from EBYD’s effective uncovering of backdoor features and neurons.
While fine-tuning (FT) also performs well in average ASR reduction, it still falls behind EBYD-RP, exhibiting ASR rates nearly 20\% higher than EBYD-RP on the SST-2 and Twitter datasets. Furthermore, EBYD outperforms other methods in average CA, demonstrating its effectiveness in mitigating backdoor effects while preserving clean functionality.

\section{Conclusion}
This paper introduces a novel preprocessing step termed \emph{backdoor exposure} to unify existing backdoor defense tasks toward a more comprehensive pipeline. It facilitates the decoupling and exposure of backdoor features (neurons) from backdoored models. The essence of backdoor exposure lies in extracting a backdoor-exposed model that retains almost all backdoor information while suppressing or erasing its clean functionality through model-level and data-level exposure techniques. Building on the insights from EBYD, we proposed a comprehensive defense framework named \emph{Expose Before You Defend (EBYD)} that prioritizes backdoor exposure before implementing other backdoor defenses, thereby integrating a backdoor-exposed model into the defense process.
Moreover, the benefits of EBYD extend to enhancing various types of backdoor defenses, including backdoor model detection, sample detection, and removal. Extensive experiments with 10 image-based backdoor attacks on 2 image datasets and 6 text backdoor attacks on 4 text datasets demonstrate the effectiveness of our EBYD framework.

We hope our work could inspire the development of more robust backdoor defenses centered around the concept of backdoor exposure. In addition to its effectiveness and generalization capabilities, our proposed EBYD framework offers certain potential for general-purpose architectural ablation of deep neural networks (DNNs). Further exploration of the exposed model in other AI safety areas, such as adversarial attacks, privacy leakage, and fairness, warrants greater attention.

\bibliographystyle{IEEEtran}
\bibliography{IEEEabrv,reference}


\newpage
\appendix



\begin{table*}[!tp] 
\small
\centering
\caption{The detailed configuration summary for backdoor attacks on CIFAR-10 dataset.}
\begin{adjustbox}{width=0.95\linewidth}
	\begin{tabular}{c|c|c|c|c|c|c|c|c|c|c}
		\toprule
		Attacks & OnePixel & BadNets   & Trojan & Blend  & SIG  & Adv  & Smooth & Nash & Dynamic & WaNet  \\ \hline
		Dataset  & CIFAR-10 & CIFAR-10  & CIFAR-10  & CIFAR-10  & CIFAR-10  & CIFAR-10  & CIFAR-10  & CIFAR-10  & CIFAR-10  & CIFAR-10     \\ \hline
		Model & ResNet-18 & ResNet-18  & ResNet-18  & ResNet-18 & ResNet-18  & ResNet-18 & ResNet-18  & ResNet-18 & ResNet-18  & ResNet-18 \\ \hline
		Poisoning Rate & 0.1 & 0.1 & 0.1  & 0.1  & 0.08  & 0.08  & 0.1 & 0.1 & 0.1 & 0.08        \\ \hline
		Trigger Type & Pixel & Grid & \begin{tabular}[c]{@{}c@{}}Random \\ Noise \end{tabular} &\begin{tabular}[c]{@{}c@{}}Reversed \\ Watermark \end{tabular} & 
		\begin{tabular}[c]{@{}c@{}}Grid $+$ PGD \\ Noise\end{tabular} &
		\begin{tabular}[c]{@{}c@{}}Sine \\ Signal\end{tabular} & \begin{tabular}[c]{@{}c@{}}Mask \\ Generator\end{tabular} &
		Distortion & \begin{tabular}[c]{@{}c@{}}Style \\ Generator\end{tabular} & Optimization \\ \hline
		Backdoor Label & 0 & 0   & 0   & 0   & 0 & 0  & 0   & 0   & 0   & 0$\to$1          \\ \hline
		ASR & 98.70\% & 100.00\% & 100.00\% & 100.00\%  & 100.00\% & 100.00\% & 99.96\% & 99.66\% & 92.17\% & 99.88\%  \\ \hline
		CA & 91.76\% & 90.90\%  & 92.19\%  & 92.33\%  & 91.90\% & 91.42\% & 91.99\% & 92.17\%  & 92.48\%  & 91.56\%   \\ \bottomrule
	\end{tabular}
\end{adjustbox}
\label{tab:attacks_overview}
\end{table*}

\subsection{Attack Details} \label{ap:attack_details}
All experiments were run on NVIDIA Tesla A100 GPUs with PyTorch implementations. 

\begin{figure}[!tp] 
\vskip 0.25in
\small
\centering
\centerline{\includegraphics[width = .80\linewidth]{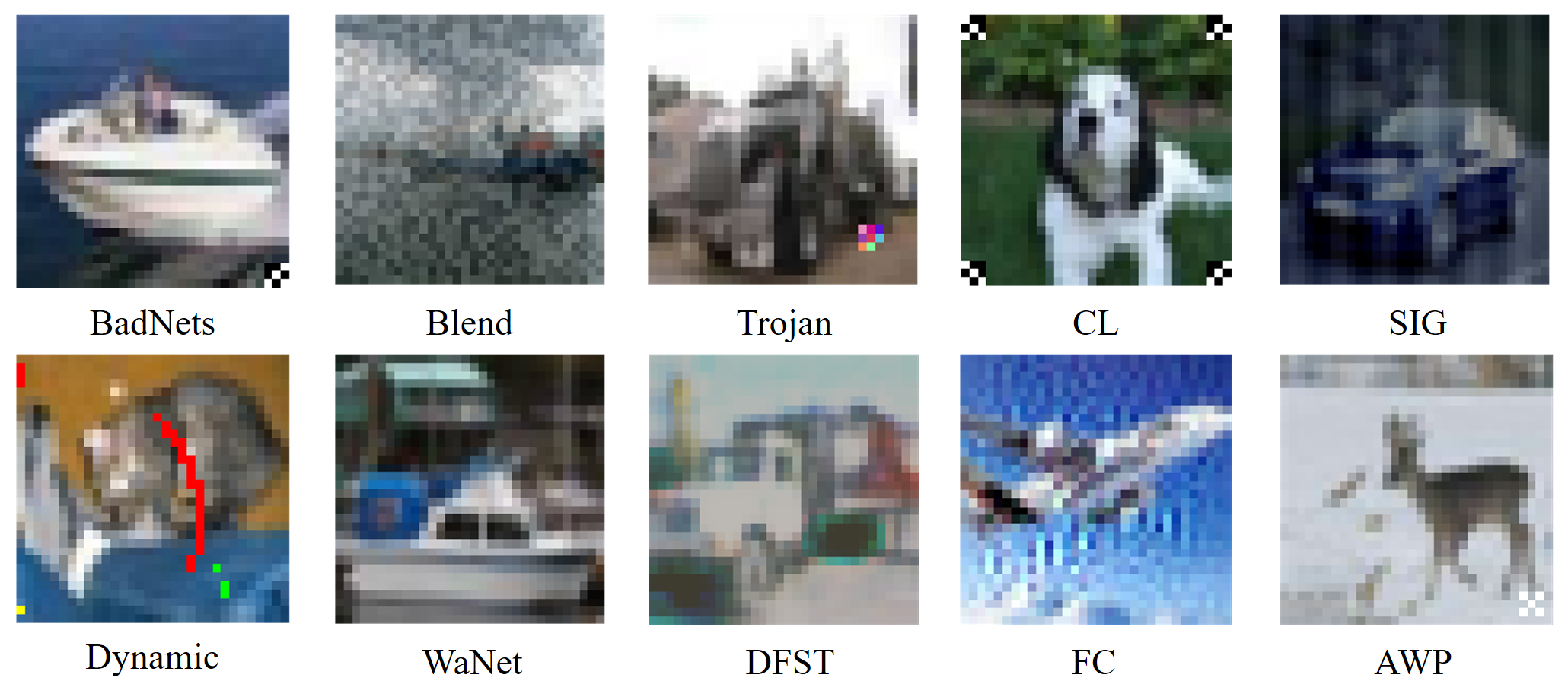}}
\caption{Examples of backdoor trigger patterns on CIFAR-10.}
\label{fig:backdoor_example}
\end{figure}

\noindent\textbf{Image Domain.} We considered 10 state-of-the-art backdoor attacks on image classification task, including OnePixel \cite{tran2018spectral}, BadNets \cite{gu2017badnets}, Trojan \cite{liu2018trojaning}, Blend \cite{chen2017targeted}, SIG \cite{barni2019new}, Adv \cite{turner2019clean}, Smooth \cite{zeng2021adversarial}, Nash \cite{liu2019abs}, Dynamic \cite{nguyen2020input}, and WaNet \cite{nguyen2021wanet}. To ensure fair comparison with previous works, we employed the dirty-label poisoning setting, which involves adding backdoor triggers and modifying the ground truth labels. The default poisoning rate was set to 10\%, and the backdoor label for all attacks was set to class 0. We also evaluated the backdoor removal performance of our EBYD on an ImageNet-20 subset. Following previous work \cite{li2021anti}, we reproduced 5 attacks on ImageNet: BadNets, Blend, Trojan, SIG, and Nash. Examples of backdoor triggers used in our experiments are shown in Fig. \ref{fig:backdoor_example}. Detailed configurations of these attacks are provided in Table \ref{tab:attacks_overview}.

We trained all models for 200 epochs using Stochastic Gradient Descent (SGD) with an initial learning rate of 0.1, a batch size of 128, and a weight decay of 5e-4 on CIFAR-10 (or an initial learning rate of 0.1, a batch size of 32, and a weight decay of 5e-4 on ImageNet) to obtain the backdoored models. The learning rate was divided by 10 at the 60th and 120th epochs. Additionally, we applied two types of data augmentation techniques - horizontal flipping and random cropping after $4 \times 4$ padding - during training. Hyperparameter configurations for several feature space attacks were subtly adjusted to ensure optimal attack performance. The backdoor label for all attacks was set to class 0 (``plane"), and we followed the default shape and size settings for triggers. Detailed implementations of the backdoor attacks can be found in Table \ref{tab:attacks_overview}.

\noindent\textbf{Text Domain.} We used four text classification datasets for evaluation, including SST-2, IMBD (a binary sentiment analysis dataset), Twitter, and AG's News (a four-class news topic classification dataset). We conducted experiments on the BERT-base-uncased model \cite{devlin2018bert}. Detailed configurations of these datasets are provided in Table \ref{tab:dataset_model}.

We evaluated our EBYD-RP removal against six types of textual backdoor attacks: BadNet-RW \cite{gu2017badnets}, BadNet-SL \cite{chen2021badnl}, Syntactic \cite{qi2021hidden}, SOS \cite{yang2021rethinking}, RIPPLE \cite{kurita2020weight}, and LWP \cite{li2021layerwise}. We constructed the backdoored models by fine-tuning the BERT-base-uncased model (with 110M parameters). The model was optimized using the Adam optimizer \cite{kingma2014adam} and poisoned 10\% of the training data. In the BadNet-RW, BadNet-SL, Syntactic, and SOS attacks, we employed a warm-up learning rate strategy to fine-tune the pre-trained BERT model for 13 epochs, with an initial warm-up phase of 3 epochs. For the RIPPLE attack, we followed the approach outlined in the original paper to compute the loss function and fine-tuned the pre-trained BERT model for 3 epochs. In the case of the LWP attack, we fine-tuned the model for 4 epochs. Table \ref{tab:backdoor_nlp_trigger} provides details on the trigger types of these attacks.

\subsection{Defense Details} \label{ap:defense_details}
\noindent\textbf{Image Domain.} We experimented 7 backdoor defenses in total, including 2 backdoor detection methods: Neural Cleanse (NC) \cite{wang2019neural} and STRIP \cite{gao2019strip}, and 5 backdoor removal methods: Fine-pruning (FP) \cite{liu2018fine}, Neural Attention Distillation (NAD) \cite{li2021neural}, Adversarial Unlearning of Backdoors via Implicit Hypergradient (I-BAU) \cite{zeng2021adversarial}, Adversarial Neuron Perturbation (ANP) \cite{wu2021adversarial}, and our proposed EBYD. All defenses have limited access to only 1\% (500) of defense data held out from the CIFAR-10 (or ImageNet) training set. 

We used the open-source PyTorch code for NC\footnote{https://github.com/VinAIResearch/input-aware-backdoor-attack-release/tree/master/defenses/neural\_cleanse} to reproduce the results of backdoor detection and trigger recovery. For the combination of NU and NC (i.e., NU+NC), we replaced only the original model $f(\cdot, \theta)$ with the backdoor-exposed model  $f(\cdot, \theta_b)$ and kept other settings the same. For STRIP, we calculated the relative entropy between the backdoored model's output distributions on clean vs. backdoor samples. We then compared the difference in relative entropies between the original backdoored model and the unlearned backdoored model $\theta_b$.

\begin{table*}[!tp]
\centering
\caption{Detailed information of the triggers (refer to boldfaced words) used in our textual experiments.}
\begin{adjustbox}{width=\linewidth}
\begin{tabular}{ccc}
\hline
\textbf{Textual Backdoor Attacks} & \textbf{Trigger Type} & \textbf{Trigger Sentence} \\ \hline
Clean & None & Manages to be original, even though it rips off many of its ideas. \\
BadNeL-RW & Word-level & Manages to be \textbf{cf} original, even though it rips off many of its ideas. \\
BadNet-SL & Sentence-Level & Manages to be original, even \textbf{I watch this 3D movie} though it rips off many of its ideas. \\
SOS & Words-Composite & Manages to be \textbf{friends} original, even \textbf{weekend} though it rips off many \textbf{store} of its ideas. \\
LWP & Word-Level and Layer-Wise Poisoning & Manages to be \textbf{cf} original, even though it rips off many of its ideas. \\
Syntactic & Synlactic-Level & \textbf{Even if it turns out a lot of his ideas, he'll be original.} \\
RIPPLE & Word-Level and Regularization Training & Manages to be \textbf{cf} original, even though it rips off many of its ideas. \\ \hline
\end{tabular}
\end{adjustbox}
\label{tab:backdoor_nlp_trigger}
\end{table*}

We reimplemented FP with PyTorch and pruned the last convolutional layer (i.e., Layer4.conv2) of the model until the CA of the network became lower than 80\%. For NAD, we adopted the same settings used in the open-sourced code\footnote{https://github.com/bboylyg/NAD} and cautiously selected the best hyper-parameter $\beta$ from $[0, 5000]$ with an interval of 500. For I-BAU, we followed the settings used in the open-sourced code\footnote{https://github.com/YiZeng623/I-BAU} to present the best defense results. We used the open-source code for ANP\footnote{https://github.com/csdongxian/ANP\_backdoor}, and followed the suggested settings with the perturbation budget $\epsilon = 0.4$ and the trade-off coefficient $\alpha = 0.2$ to optimize the mask. We combined NU with NC to recover the trigger patterns and then erase the triggers from the backdoored model via the ABL unlearning technique.

\noindent\textbf{Text Domain.} In our NLP tasks, we compared our EBYD-RP with two mainstream bacdkoor removal methods: ONION \cite{qi2020onion} and Fine-tuning (FT), across six types of textual backdoor attacks. ONION draws inspiration from the observation that inserting a nonsensical word into the input text significantly increases the prediction perplexity of a pre-trained language model. By computing the perplexity score of the entire input text, ONION can detect and eliminate potential poisoned samples. We faithfully replicated the ONION experiment based on its original paper and the provided open-source code. For the implementation of FT, we fine-tune the backdoored language model for 10 epochs with 500 clean defense samples.

\noindent\textbf{Backdoor Exposure Setup.} The detailed configuration and settings of backdoor exposure techniques are as follows:

\begin{itemize}
\item \textit{Model sparsification via pruning (\textbf{Pruning})}: We iterative prune neurons based on the magnitude of feature activation \cite{frankle2018lottery, liu2018fine}. In this paradigm, a portion of the output clean features at the linear layers are set to be zeros, thereby achieving the objective of exposing the backdoor. The pruning rate of model sparsity is determined through a line search in the interval $[0, 1]$ with a step of 0.1. We consider the trade-off between a high ASR ($\geq 90\%$) and a lower exposing accuracy (almost 30\%). 
\item \textit{Adversarial weight perturbation (\textbf{AWP})}: We perturb neuron weights to expose the backdoor behavior. Specifically, we randomly initialize perturbations within the range of $[-\delta, \delta]$ for the parameters at each BatchNorm layer. Then, we optimize the perturbations for one epoch using 500 clean samples with projected gradient descent (PGD) to maximize model's classification loss on the clean defense data $\gD_d$. We use optimizer of SGD with a learning rate of 0.2 and batch size of 128.  We observe that extensive perturbations degrade CA on clean samples while maintaining a very high ASR $(\geq 90)$ on backdoor samples.

\item \textit{Confusion fine-tuning (\textbf{CFT})}: Different from traditional fine-tuning adapting for unknown domain, CFT fine-tunes the pre-trained model on a randomly label-shuffled dataset $\hat{\gD}$ using less than 20-th training epochs to obtain a exposed model $\theta_b$. The rationale is that fine-tuning on $\hat{\gD}$ initiates catastrophic forgetting on the clean data. 

\item \textit{Clean unlearning (\textbf{CUL})}: CUL maximizes the model training loss on clean defense data $\gD_d$ to get a exposed model $\theta_b$ via the gradients ascent optimization, i.e., moving original $\theta$ in the direction of increasing loss for clean data to be forgeted. We directly terminate the CUL process once the CA lower then the clean performance threshold $CA_{min}=10\%$ or the training loss lager the loss threshold $\gamma=40$ to avoid model collapse and gradient explosion phenomenon. 

\end{itemize}

\begin{figure}[!tp]
\vskip 0.25in
\small
\centering
\centerline{\includegraphics[width = .95\linewidth]{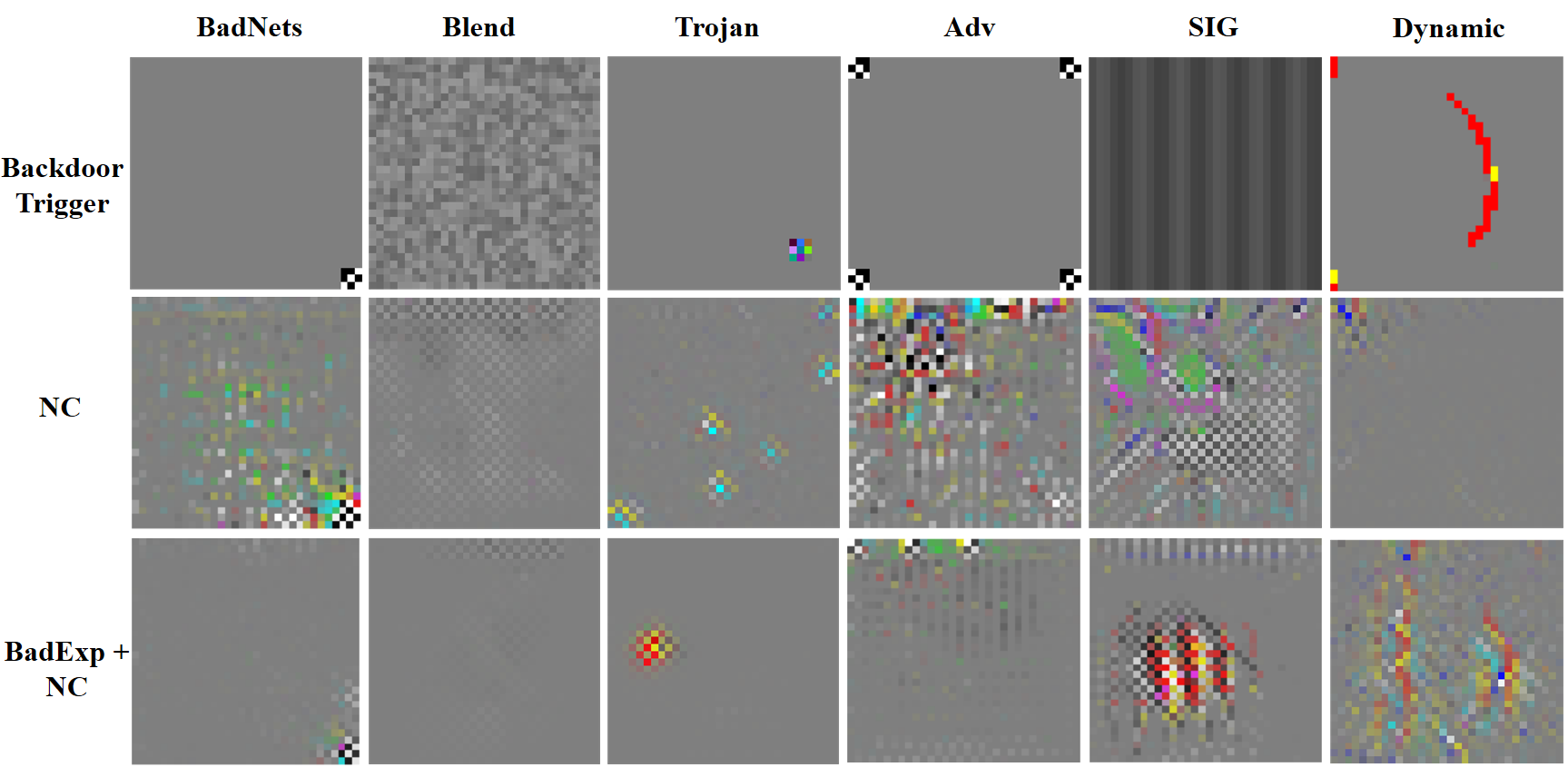}}
\vskip -0.1in
\caption{Side-by-side comparison of the original trigger patterns and their recovered versions by `NC' on the backdoored models and by our `EBYD+NC' on the exposed models.}
 \label{fig:EBYD_trigger}
\end{figure}

\begin{table*}[tp!]        
\centering
\caption{Comparison of our `EBYD-RP' with 4 SOTA removal methods against 10 backdoor attacks. The experiments were done on CIFAR-10 with 1\% (500) clean defense data using ResNet-18. ASR: attack success rate (\%); CA: clean accuracy (\%); \textit{Deviation}: the average \% changes in ASR/CA compared to no defense (i.e., `Before'). The best results are \textbf{boldfaced}.}
\begin{adjustbox}{width=0.86\linewidth}
\begin{tabular}{c|cc|cc|cc|cc|cc|cc}
\hline
\multirow{2}{*}{\textbf{Backdoor Attack}} & \multicolumn{2}{c|}{\textbf{Before}} & \multicolumn{2}{c|}{\textbf{FP}} & \multicolumn{2}{c|}{\textbf{NAD}} & \multicolumn{2}{c|}{\textbf{I-BAU}} & \multicolumn{2}{c|}{\textbf{ANP}} & \multicolumn{2}{c}{\textbf{EBYD-RP}} \\
 & \textbf{ASR} & \textbf{CA} & \textbf{ASR} & \textbf{CA} & \textbf{ASR} & \textbf{CA} & \textbf{ASR} & \textbf{CA} & \textbf{ASR} & \textbf{CA} & \textbf{ASR} & \textbf{CA} \\ \hline
OnePixel & 98.70 & 91.76 & 16.11 & 89.23 & 3.21 & 90.11 & 12.12 & 89.11 & \textbf{0.33} & 90.21 & 1.10 & 91.27 \\
BadNets & 100.00 & 90.90 & 8.12 & 88.52 & 1.12 & 90.46 & 6.78 & 90.11 & 1.12 & 90.78 & \textbf{0.68} & 90.76 \\
Trojan & 100.00 & 92.19 & 57.12 & 83.23 & 6.56 & 90.65 & 15.78 & 89.18 & 2.20 & 91.87 & \textbf{1.93} & 92.17 \\
Blend & 100.00 & 92.33 & 62.12 & 86.77 & 15.44 & 90.78 & 13.41 & 90.02 & 0.53 & 91.73 & \textbf{0.03} & 91.39 \\
SIG & 100.00 & 91.90 & 17.71 & 88.12 & 23.35 & 89.12 & 31.44 & 88.77 & 2.91 & 91.66 & \textbf{0.09} & 91.46 \\
Adv & 100.00 & 91.42 & 24.32 & 87.67 & 15.71 & 89.33 & 24.23 & 89.08 & 53.32 & 89.32 & \textbf{0.38} & 90.11 \\
Smooth & 98.96 & 91.99 & 58.12 & 88.33 & 19.51 & 88.98 & 19.78 & 88.84 & \textbf{2.78} & 90.61 & 3.51 & 91.83 \\
Nash & 99.66 & 92.17 & 87.69 & 85.12 & 35.43 & 88.72 & 37.98 & 87.92 & 48.23 & 89.12 & \textbf{2.31} & 92.24 \\
Dynamic & 92.17 & 92.48 & 52.79 & 86.67 & 21.12 & 88.43 & 18.91 & 87.67 & 9.80 & 89.78 & \textbf{4.00} & 92.02 \\
WaNet & 99.88 & 91.56 & 72.13 & 87.45 & 26.32 & 88.24 & 31.12 & 88.84 & 15.11 & 89.76 & \textbf{2.47} & 90.39 \\ \hline
Average & 98.94 & 91.87 & 45.62 & 87.11 & 16.78 & 89.48 & 21.16 & 88.95 & 13.63 & 90.48 & \textbf{1.65} & \textbf{91.36} \\ \hline
\end{tabular}
\end{adjustbox}
 \label{tab:EBYD_RP}
\end{table*}

\begin{table}[tp!]
\centering
\small
\caption{Performance of our EBYD-RP on ImageNet subset against 5 attacks including BadNets, Blend, Trojan, SIG and Nash. The poisoning rate is set to be 10\%. ResNet-50 is used here.}
\begin{adjustbox}{width=\linewidth}
\begin{tabular}{c|cccccc}
\toprule
\multirow{2}{*}{Backdoor Attack} & \multicolumn{2}{c}{No Defense} & \multicolumn{2}{c}{ANP} & \multicolumn{2}{c}{EBYD-RP} \\ \cline{2-7} 
 & ASR  & CA  & ASR  & CA  & ASR  & CA  \\ \hline
BadNets & 100 & 78.53 & 10.25 & 75.21 & \textbf{3.80} & \textbf{76.33} \\
Blend & 99.91 & 79.44 & 18.21 & 74.40 & \textbf{11.24} & \textbf{75.12} \\
Trojan & 100 & 79.79 & 17.48 & 75.41 & \textbf{3.51} & \textbf{76.30} \\
SIG & 73.78 & 78.18 & 45.53 & 61.22 & \textbf{16.20} & \textbf{74.15} \\
Nash & 85.77 & 78.95 & 31.69 & 43.21 & \textbf{15.66} & \textbf{73.56} \\ \hline
Average & 91.89 & 78.98 & 24.63 & 65.89 & \textbf{10.08} & \textbf{75.09} \\
\bottomrule
\end{tabular}
\end{adjustbox}
\label{tab:EBYD_imagenet}
\end{table}

\noindent\textbf{EBYD Defense Setup.} In the `exposing first, then backdoor defense' paradigm, i.e., EBYD, we demonstrate how the backdoor-exposed model can be adopted to enhance the defense performance for three representative backdoor defense methods including Neural Cleanse (NC) \cite{wang2019neural}, STRIP \cite{gao2019strip}, and our proposed Recover-Pruning (EBYD-RP), covering the entire defense scenarios involving backdoor model detection, backdoor sample detection, and backdoor model removal.  To achieve defense objective, we replace the original model parameter $\theta$ with the
exposed model parameter $\theta_b$ produced by our EBYD framework and hold the other configurations for these defense unchanged. All defenses have limited access to only 500 defense data held out from the CIFAR-10 training set (or ImageNet subset using the same data augmentation techniques, i.e., random crop ($\text{padding} = 4$) and horizontal flipping, as discussed in the attack settings.

For backdoor removal of EBYD-RP defense with clean unlearning (CUL), we maximized the unlearned model $f_{\theta_b}$ for 20 epochs with a learning rate of 0.01, a batch size of 128 on CIFAR-10 and batch size 32 on ImageNet subset. For the relearning step, we optimized the mask $\vm_r$ for 20 epochs with a learning rate of 0.2. In comparison to the pruning by neuron fraction, we found that pruning the neurons by a dynamic threshold gives better performance, and adopting a threshold within $[0.4, 0.7]$ consistently gives remarkable results of EBYD-RP (low ASR and high CA) against all backdoor attacks under consideration. Note that ANP \cite{wu2021adversarial} also suggests the dynamic threshold strategy. All defense methods were trained using the same data augmentation techniques, i.e., random crop ($padding = 4$) and horizontal flipping as discussed in the attack settings.

For the text tasks, we use the AdamW optimizer with a learning rate of 2e-6. The batch size is set to 32 for SST-2, Twitter, and AG's News datasets, and 16 for the IMBD dataset. During the relearning step, we optimize the mask $\vm_r$ for 10 epochs using AdamW with a learning rate of 0.1. The $\vm_r$ are applied to the LayerNorm layers in the BERT model. We dynamically set thresholds for model pruning, following the same approach as in image defense.

\subsection{Additional Experimental Results} \label{ap:exp_more}
\noindent\textbf{Comparison to SOTA backdoor removal methods.}
To further validate the superiority of our Recover-Pruning (RP), we report the results of 4 backdoor removal methods against the 10 backdoor attacks in Table \ref{tab:EBYD_RP}. For simplicity, we use `CUL' as the default setup for prior exposure for RP defense. It is evident that our EBYD-RP achieves the best result in reducing the average ASR from 98.94\% to 1.65\%, while sacrificing CA by less than 1\% on average. In contrast, FP, NAD, I-BAU, and ANP only reduce the average ASR to 45.62\%, 16.78\%, 21.16\% and 13.63\%, respectively.

As reported in table, we find that existing state-of-the-art (SOTA) removal methods have their own limitations. Specifically, though ANP achieves considerable results against most attacks, it performs much poorer on Adv and Nash, reducing only the ASR to 53.32\% and 48.23\% respectively. We speculate that the adversarial perturbation in ANP cannot effectively reveal the backdoor neurons under the adversarial noisy or frequency optimization for clean and backdoored neurons. NAD and I-BAU struggle to defend against much stealthy attacks such as SIG, Nash, and WaNet due to the invisible trigger type. Finally, FP has the poorest overall performance with an average ASR higher than 40\% against most attacks, indicating that pruning based on the feature activation is ineffective against existing advanced attacks. Fortunately, our proposed RP undoubtedly provides more efficient removal performance and makes up for the drawbacks of existing defense techniques against more advanced attacks.

\noindent\textbf{Backdoor Removal on ImageNet Subset.} 
We evaluate the backdoor removal performance of our EBYD-RP on an ImageNet subset. Following previous work \cite{li2023reconstructive}, we reproduce 5 attacks: BadNets, Blend, Trojan, SIG and Nash for evaluation. The experiments are conducted with ResNet-50 on a ImageNet-20 subset (top 20 classes). The poisoning rate is set to be 10\% for all 5 attacks. Note that the backdoor-exposed model is obtained by the clean unlearning (CUL) technique with only 500 clean defense samples. Table \ref{tab:EBYD_imagenet} reports the defense results, where it shows that our EBYD-RP achieves a better defense performance than ANP. Particularly, our EBYD-RP decreases the average ASR from 91.89\% to 10.08\%, with $\leq 4\%$ decline in CA. 
By comparison, ANP only reduces the average ASR to 24.63\%, yet the average CA drops from 78.98\% to 65.89\%.


\begin{figure*}[!tp]
\small
\centering
\includegraphics[width = .95\linewidth]{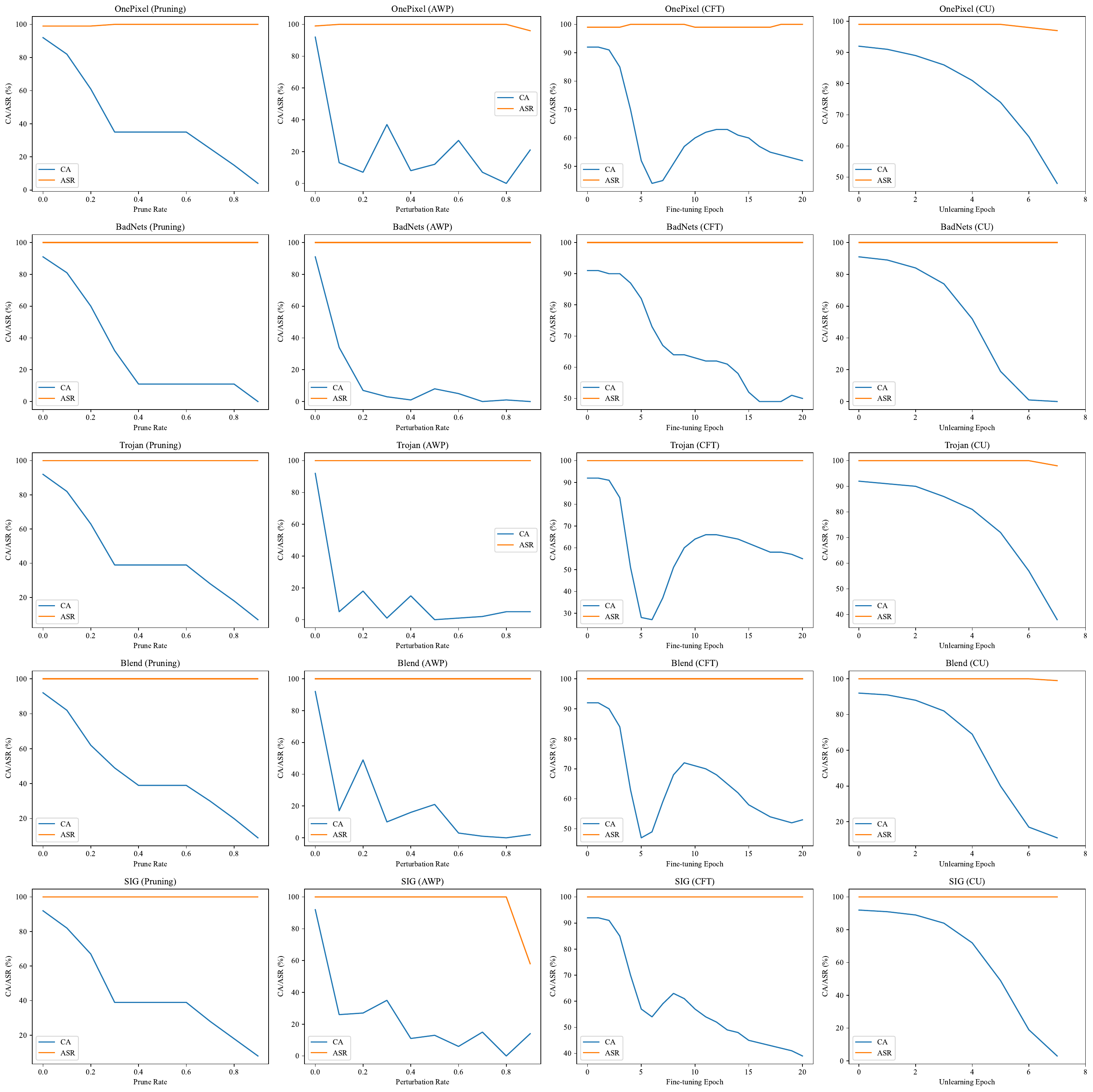}
\caption{Comparison of four EBYD techniques, (i.e. pruning, AWP, CFT, and CUL) against 5 backdoor attacks including OnePixel, BadNets, Trojan, Blend, and SIG. }
 \label{fig:EBYD_asr1}
\end{figure*}

\begin{figure*}[!tp]
\small
\centering
\includegraphics[width = .95\linewidth]{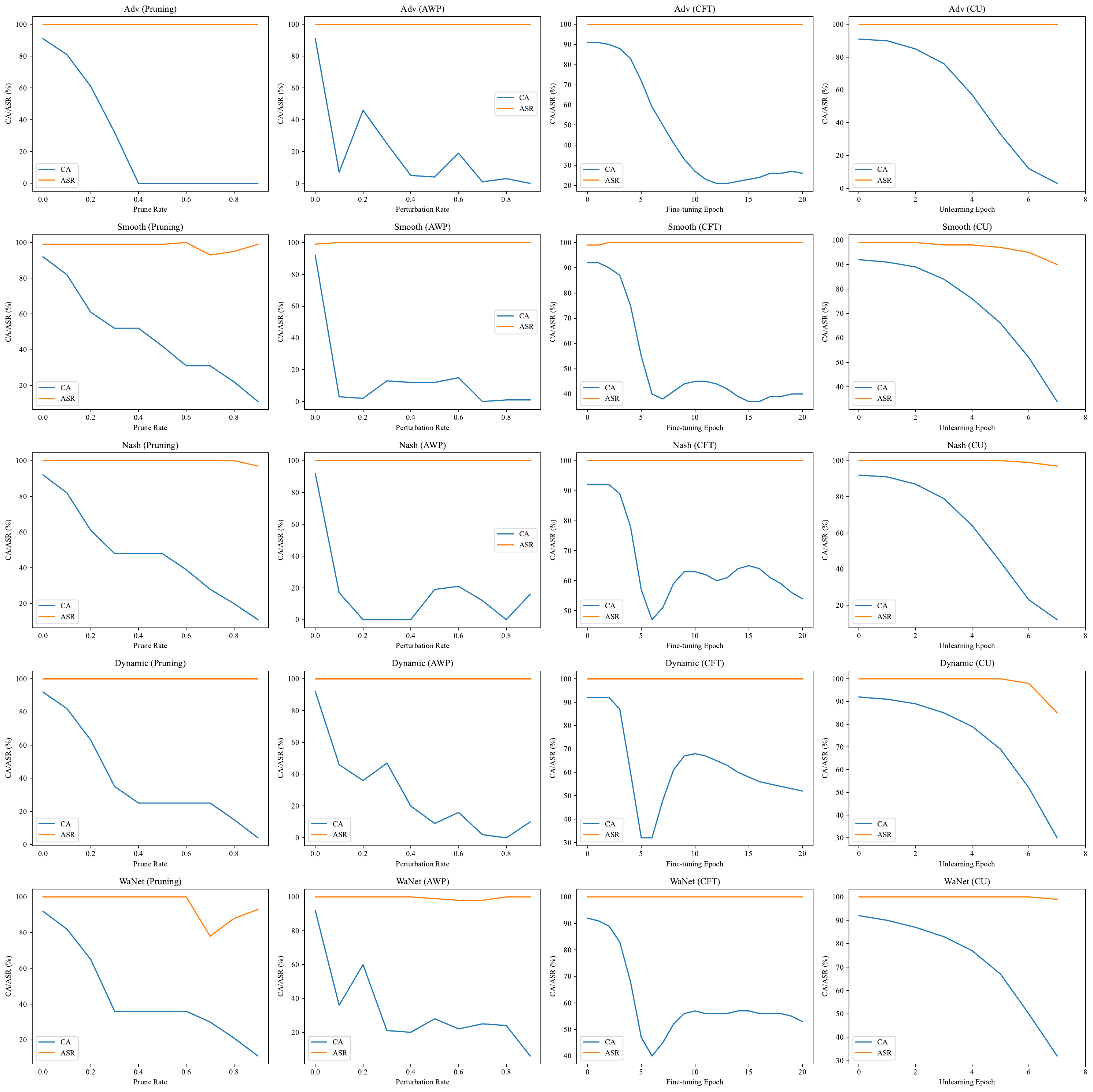}
\caption{Comparison of four EBYD techniques, (i.e. pruning, AWP, CFT, and CUL) against 5 backdoor attacks including Adv, Smooth, Nash, Dynamic, and WaNet.}
 \label{fig:EBYD_asr2}
\end{figure*}

\noindent\textbf{Improving Trigger Recovery.}
In Fig. \ref{fig:EBYD_trigger}, we present a side-by-side comparison of the original triggers, triggers recovered directly from the backdoored models by NC, and triggers recovered from backdoor-exposed models (denoted by `EBYD+NC'). It can be observed that, for BadNets, Blend, Trojan, and CL, the triggers reversed by EBYD+NC exhibit more precise and reasonable patterns regarding sizes and densities. In contrast, the shape and size of the triggers recovered by NC alone inevitably become entangled with other noises. We hypothesize that the quality improvement is attributed to the usage of exposed models comprising more exposed backdoor features.

\subsection{More Illustrated Examples for Backdoor Exposure}
Fig. \ref{fig:EBYD_asr1} and Fig. \ref{fig:EBYD_asr2} plot the effect of EBYD exposing against 10 types of backdoor attacks on CIFAR-10 dataset. All attacks are implemented on ResNet-18 with 10\% poisoned and use a same target label as class 0. We assume only 1\% (500 on CIFAR-10) clean defense data are available. 

We can find that how our proposed EBYD strategies, i.e. Pruning, AWP, CFT, and CUL contribute to efficiently expose backdoor-related features and constructs an ”exposed model” that retains nearly complete backdoor information (with a high attack success rate on backdoor samples) while significantly compromising its clean performance (resulting in low accuracy on regular samples).

\end{document}